\PassOptionsToPackage{table}{xcolor}
\documentclass[sn-vancouver,Numbered]{sn-jnl}

\usepackage{graphicx}%
\usepackage[title]{appendix}%
\usepackage{manyfoot}%
\usepackage[utf8]{inputenc}
\usepackage{tikz}
\usepackage{float}
\usetikzlibrary{tikzmark,arrows.meta,calc,positioning}
\usepackage[table]{xcolor}
\usepackage{tcolorbox}
\usepackage{soul}
\usepackage{amsmath}
\usepackage{pifont}
\usepackage{pdfrender}
\usepackage{placeins}
\usepackage{setspace}
\usepackage{geometry}
\usepackage{comment}
\usepackage{todonotes}
\usepackage{lmodern}
\usepackage{makecell}
\definecolor{mylightblue}{RGB}{0,191,235}
\definecolor{anotherlightblue}{RGB}{204,229,255}
\sethlcolor{anotherlightblue}
\definecolor{anothergreen}{RGB}{217,255,217} 
\sethlcolor{anothergreen}

\begin{document}



\title{Unleashing the potential of prompt engineering for large language models}

\author[1]{\fnm{Banghao} \sur{Chen}}
\email{chenbanghao@u.nus.edu}
\author[1]{\fnm{Zhaofeng} \sur{Zhang}}
\email{zhangzf@umich.edu}
\author*[1]{\fnm{Nicolas} \sur{Langrené}}\email{nicolaslangrene@uic.edu.cn}
\author*[2,1]{\fnm{Shengxin} \sur{Zhu}}\email{shengxin.zhu@bnu.edu.cn}

\date{\vspace{0.25em}
First version: October 23, 2023\\
\vspace{0em}
This revised version: \today\\
}

\affil[1]{Guangdong Provincial Key Laboratory of Interdisciplinary Research and Application for Data Science, \orgname{BNU-HKBU United International College}, \orgaddress{\city{Zhuhai} \postcode{519087}, \state{Guangdong}, \country{China}}}
\affil[2]{\orgname{Research Center for Mathematics, Beijing Normal University}, \orgaddress{\street{No.18, Jingfeng Road}, \city{Zhuhai} \postcode{519087}, \state{Guangdong}, \country{China}}}

\abstract{This comprehensive review delves into the pivotal role of prompt engineering in unleashing the capabilities of Large Language Models (LLMs). The development of Artificial Intelligence (AI), from its inception in the 1950s to the emergence of advanced neural networks and deep learning architectures, has made a breakthrough in LLMs, with models such as GPT-4o and Claude-3, and in Vision-Language Models (VLMs), with models such as CLIP and ALIGN. Prompt engineering is the process of structuring inputs, which has emerged as a crucial technique to maximize the utility and accuracy of these models. This paper explores both foundational and advanced methodologies of prompt engineering, including techniques such as self-consistency, chain-of-thought, and generated knowledge, which significantly enhance model performance. Additionally, it examines the prompt method of VLMs through innovative approaches such as Context Optimization (CoOp), Conditional Context Optimization (CoCoOp), and Multimodal Prompt Learning (MaPLe). Critical to this discussion is the aspect of AI security, particularly adversarial attacks that exploit vulnerabilities in prompt engineering. Strategies to mitigate these risks and enhance model robustness are thoroughly reviewed. The evaluation of prompt methods is also addressed, through both subjective and objective metrics, ensuring a robust analysis of their efficacy. This review also reflects the essential role of prompt engineering in advancing AI capabilities, providing a structured framework for future research and application.}


\maketitle


\newpage

\section{Introduction}
In recent years, a significant milestone in artificial intelligence research has been the progression of natural language processing capabilities, primarily attributed to Large Language Models (LLMs). Many popular models, rooted in the transformer architecture \cite{10.5555/3295222.3295349}, undergo training on extensive datasets derived from web-based text. Central to their design is a self-supervised learning objective, which focuses on predicting subsequent words in incomplete sentences. Those models are called Artificial Intelligence-Generated Content (AIGC), and their ability to generate coherent and contextually relevant responses is a result of this training process, where they learn to associate words and phrases with their typical contexts.

LLMs operate by encoding the input text into a high-dimensional vector space, where semantic relationships between words and phrases are preserved. The model then decodes this representation to generate a response, guided by the learned statistical patterns \cite{bender2021dangers}. The quality of the response can be influenced by various factors, including the prompt provided to the model, the model's hyperparameters, and the diversity of the training data.

These models, including LLMs such as the GPT series 
\cite{brownGPT3, openai2024gpt4technicalreport} by OpenAI, along with many others (e.g. Gemini \cite{geminiteam2024geminifamilyhighlycapable, google2024gemini} and Gemini (BARD) \cite{hulbert2023bard} by Google, Claude series by Anthropic \cite{anthropic2024claude3, claude3}, and Llama series open-source model from Meta \cite{2023llama2, dubey2024llama3herdmodels}), have revolutionized tasks ranging from information extraction to the creation of engaging content \cite{Sarkhel2023}. In parallel, the development of multimodal large models (MMLMs) has introduced the ability to process and generate not just text, but also images, audio, and other forms of data, showcasing their flexibility and effectiveness. These models integrate multiple data modalities into a single framework, demonstrating strong capabilities in tasks such as image description and visual question answering (VQA). Early MMLMs include the DALL-E series \cite{ramesh2021zeroshot, marcus2022dalle2, openai2021dalle}, which can generate images from textual descriptions, and CLIP, which can understand and relate text and image data in a unified manner \cite{Radford2021LearningTV, li2022supervision}. More powerful models such as GPT-4o by OpenAI \cite{openai_gpt4o} and Claude 3.5 Sonnet by Anthropic \cite{anthropic2024claude3, claude3} excel in multimodal tasks involving text generation and understanding, integrating natural language processing with various forms of data to perform diverse and complex tasks. While numerous advanced models are currently capable of processing audio, the majority of accessible Application Programming Interfaces (APIs) remain focused on text and vision modalities. With the gradual introduction of audio APIs, a broad expansion of research in this modality can be expected \cite{Moore2024}. The evolution of LLMs reflects significant strides in AI research, characterized by increasing model complexity, enhanced training methodologies, and broader application potentials. These advancements underline the critical role of prompt engineering in maximizing the utility and accuracy of these models, ensuring that they can effectively cater to diverse and dynamic user needs. While this survey is mainly focused on prompt engineering for LLMs, the inclusion of vision-language models (VLMs) offers a broader perspective, revealing the potential and challenges of prompt engineering in handling multimodal data. By integrating research from both types of models, we can gain a deeper understanding of the applications of prompt engineering and provide valuable insights for future research and practice.

In real applications, the prompt is the input of the model, and prompt engineering can result in significant output differences \cite{kaddour2023challenges}. Modifying both the structure (e.g., altering length, arrangement of instances) and the content (e.g., phrasing, choice of illustrations, directives) of the prompt can exert a notable influence on the model's behavior \cite{lu2022sensitivity, webson2022prompt}.

\begin{figure}[h]
  \centering
  \includegraphics[width=\linewidth]{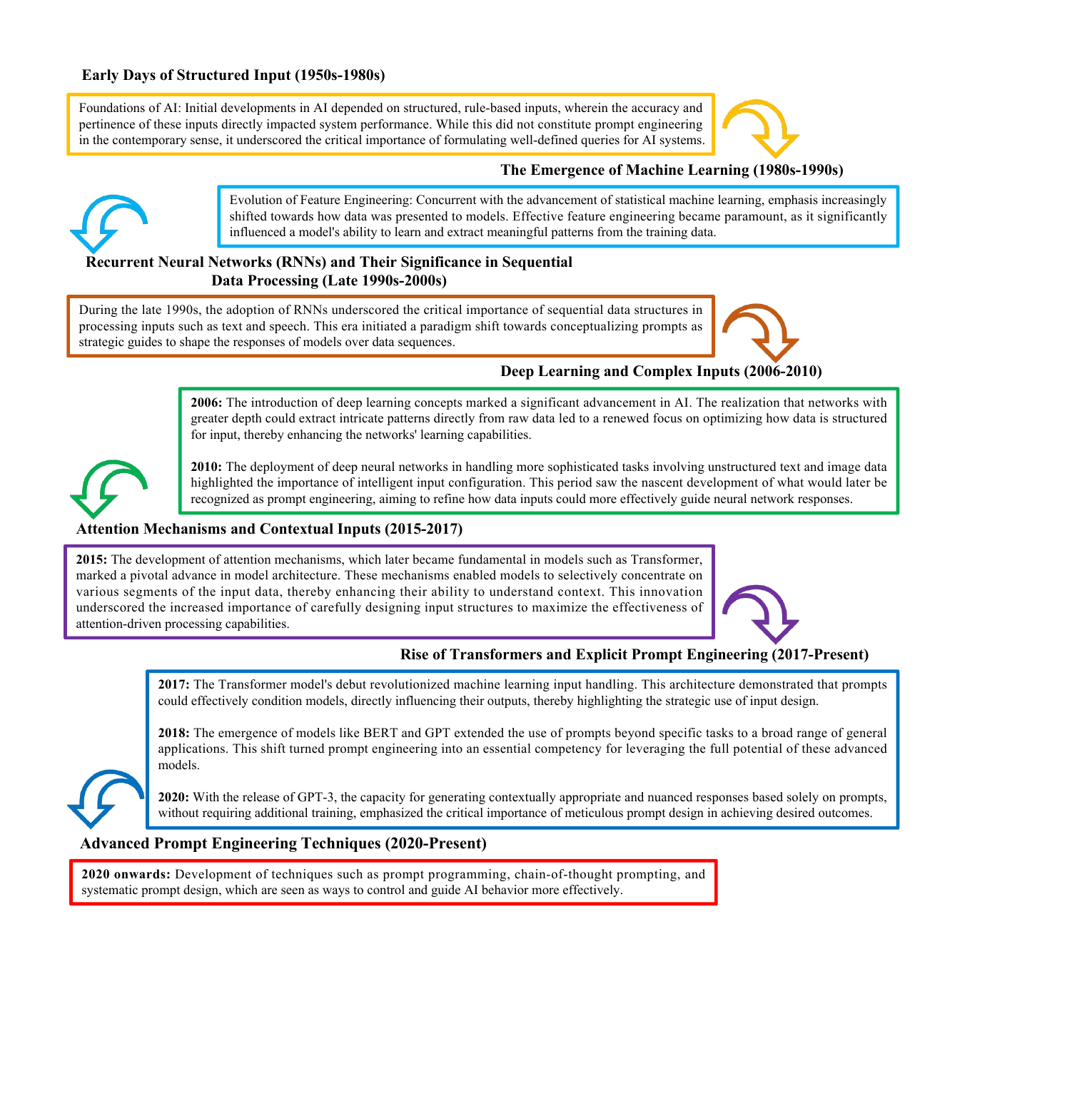}
  \caption{History of the development in prompt engineering.}
  \label{History_of_the_development_in_prompt_engineering}
\end{figure}

Prompt engineering refers to the systematic design and optimization of input prompts to guide the responses of LLMs, ensuring accuracy, relevance, and coherence in the generated output. This process is crucial in harnessing the full potential of these models, making them more accessible and applicable across diverse domains. Over time, prompt engineering has evolved from an empirical practice into a well-structured research domain. As illustrated in Figure \ref{History_of_the_development_in_prompt_engineering}, the historical progression of prompt engineering showcases significant milestones from the early days of structured inputs in the 1950s to advanced methodologies such as chain-of-thought prompting \cite{wei2022chain} and self-consistency prompting \cite{wang2023selfconsistency} developed in recent years. This review will primarily focus on techniques emerging from the period of rapid development after 2017.

Contemporary prompt engineering encompasses a spectrum of techniques, ranging from foundational approaches such as role-prompting \cite{shanahan2023roleplay} to more sophisticated methods such as chain-of-thought prompting \cite{wei2022chain}. The domain remains dynamic, with emergent research continually unveiling novel techniques and applications in prompt engineering. The importance of prompt engineering is accentuated by its ability to guide model responses, thereby amplifying the versatility and relevance of LLMs in various sectors. Importantly, a well-constructed prompt can counteract challenges such as machine hallucinations \cite{maynez2020faithfulness, bubeck2023sparks}. The influence of prompt engineering extends to numerous disciplines. For instance, it has facilitated the creation of robust feature extractors using LLMs, thereby improving their efficacy in tasks such as defect detection and classification \cite{yong2022prompt}.

This paper aims to provide a comprehensive review of the prompt engineering techniques proposed so far within the realm of LLMs. The structure of the paper is organized as follows: Section \ref{Sec2} explores the foundational methods of prompt engineering, emphasizing the importance of clear and precise instructions, role-prompting, and iterative attempts to optimize outputs. In Section \ref{sec:advanced_methodologies}, advanced methodologies such as chain-of-thought, self-consistency, and generated knowledge are introduced to guide models in generating high-quality content. Section \ref{Sec4} discusses methodologies specific to VLMs, including Context Optimization (CoOp), Conditional Context Optimization (CoCoOp), and Multimodal Prompt Learning (MaPLe), which enhance the performance of VLMs \cite{WANG2023100047}. Section \ref{Sec5} assesses the efficacy of various prompt methods through both subjective and objective evaluations, ensuring a robust analysis of their effectiveness. Section \ref{Sec6} briefly explores the applications of prompt engineering across diverse fields such as education, content creation, computer programming, and reasoning tasks, highlighting its broad impact. Section \ref{Sec7} addresses the security implications of prompt engineering, identifying common vulnerabilities in LLMs and reviewing strategies to enhance security such as adversarial training. Finally, Section \ref{Sec8}  explores prospective methodologies, emphasizing the importance of understanding AI model structures and the potential of AI agents in advancing AI-generated content tools. This structured framework provides an entire overview of the pivotal role of prompt engineering in advancing AI capabilities and guiding future research and applications.

\section{Basics of prompt engineering}
\label{Sec2}
By incorporating just a few key elements, one can craft a basic prompt that enables LLMs to produce high-quality answers. In this section, some essential components of a well-made prompt will be discussed and examples of these methods will be shown.

\subsection{Model introduction: GPT-4}
\label{sec2.1}
All of the examples in the following sections are generated by GPT-4, developed by OpenAI \cite{openai2024gpt4technicalreport}. Vast amounts of text data have been used to train GPT-4, whose number of parameters has been estimated to be several orders of magnitude larger than the 175 billion parameters that had been used for the earlier GPT-3 \cite{brownGPT3}. The architectural foundation of the model rests on transformer \cite{10.5555/3295222.3295349}, which essentially are attention mechanisms that assign varying weights to input data based on the context. Similar to GPT-3, GPT-4 was also fine-tuned to follow a broad class of written instructions by reinforcement learning from human feedback (RLHF) \cite{christiano2017deep,stiennon2020RLHF}, which is a technique that uses human preferences as a reward signal to fine-tune models. 

When GPT-4 receives an input prompt, the input text will be firstly converted into tokens that the model can interpret and process. These tokens are then managed by transformer layers \cite{Radford2019LanguageMA}, which capture their relationships and context. Within these layers, attention mechanisms distribute different weights to tokens based on their relevance and context. After attention processing, the model forms its internal renditions of the input data, known as intermediate representations. These representations are then decoded back into human-readable text \cite{radford2018improving}.

A significant aspect of this process is the randomness function \cite{Radford2019LanguageMA}. This function is influenced by two primary parameters, temperature and top-$k$ sampling. The first one, temperature \cite{holtzman2020curious} balances the randomness and determinism in the output. A higher temperature value results in more random outputs, while a lower value makes the output more deterministic. The second one, top-$k$ sampling \cite{welleck2019neural}, limits the model's choices to the top $k$ most probable tokens during each step of output generation. The final stage of this process is the output generation, where the model crafts the final text.

\subsection{Giving instructions}
\begin{figure}[h]
\begin{spacing}{1.1}
\FloatBarrier
\centering
\begin{minipage}{1\textwidth}
\large
\centering
\tikzmarknode{basic11}{}
\begin{tcolorbox}[colback=white]
\footnotesize
\raggedright
\textcolor{black}{\scriptsize I want to understand some cutting-edge aspects of technology.}\tikzmark{a}
\end{tcolorbox}
\tikzmarknode{basic12}{}
\begin{tcolorbox}[colback=white]
\scriptsize
Here are some of the key areas:\\

\hl{1. Artificial intelligence and machine learning:} These technologies are becoming increasingly sophisticated, with advancements in areas such as natural language processing (like the technology you're interacting with right now), computer vision, and reinforcement learning. Techniques such as transfer learning and federated learning are being used to make machine learning models more efficient and less data-hungry.\\
\hl{2. Quantum computing:} Quantum computers use the principles of quantum mechanics to process information. They hold the potential to solve certain types of problems much more efficiently than classical computers. Companies like Google, IBM, and many startups are investing heavily in this area.\\
\hl{3. 5G and beyond:} 5G wireless technology began widespread deployment in 2020 and 2021, bringing faster speeds, lower latency, and the ability to connect more devices simultaneously. Research is also being done on 6G, which is expected to integrate even more deeply with AI and potentially bring about more transformative changes.\\
\hl{4. Biotechnology:} ...\\
\end{tcolorbox}
\end{minipage}\hfill
\begin{tikzpicture}[overlay,remember picture,-{Latex[open, scale=0.8]},shorten >=5pt,shorten <=5pt,line width=1.5pt]
\coordinate (start) at ([xshift=0cm, yshift=-38pt]basic11.south);
\coordinate (end) at ([xshift=0cm, yshift=-18pt]$(start)!0.9!(basic12.north)$); 
\draw[white, line width=0.6mm] (start) -- (end);
\draw[black, -{Latex[open, scale=1]}] (start) -- (end);

\node[anchor=north east, draw, fill=mylightblue, text=black, minimum width=1cm, minimum height=0.4cm, inner sep=1.2pt, rounded corners] at ([xshift=-4.65cm, yshift=-0.2cm]basic11) {Model input};
\node[anchor=north east, draw, fill=green, text=black, minimum width=0.5cm, minimum height=0.2cm, inner sep=1.2pt, rounded corners] at ([xshift=-4.40cm, yshift=-0.2cm]basic12) {Model output};
\end{tikzpicture}
\caption{\scriptsize Giving instructions without extra description.}
\label{Giving Instructions without extra description.}
\end{spacing}
\end{figure}
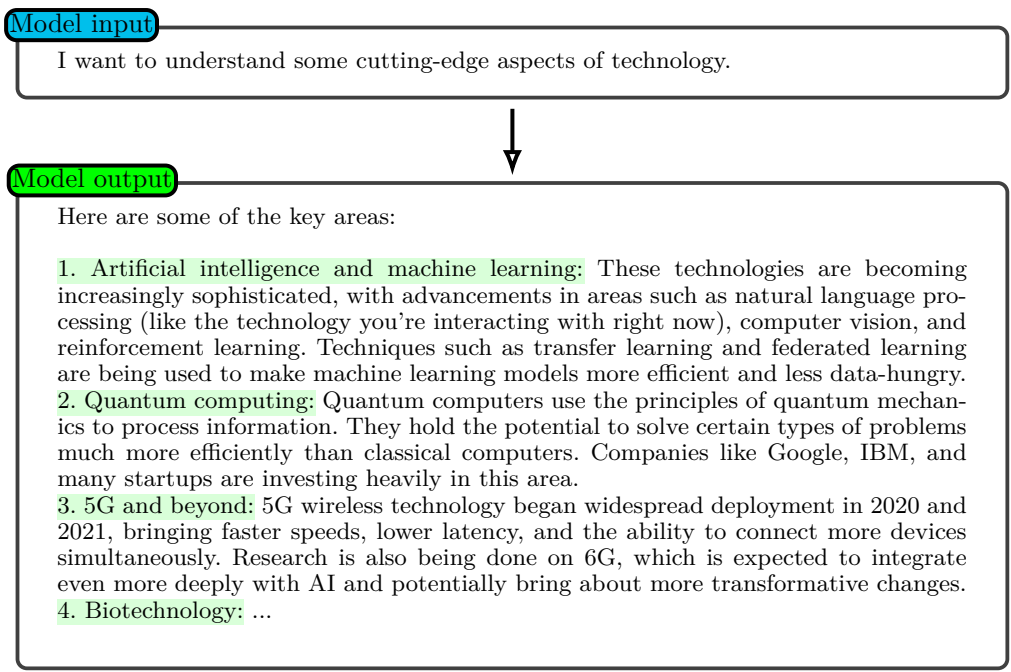

The method of giving instructions, also known as re-reading \cite{xu2023re}, refers to the heuristics of human reading strategy. It has been observed that the output generated by GPT-4, introduced in Section \ref{sec2.1}, tends to be excessively general when provided with fundamental instructions devoid of any supplementary description \cite{yansong2018joint,luo2019unsupervised}. An example prompt is shown in Figure \ref{Giving Instructions without extra description.}. When the model is prompted with a basic instruction, it faces a plethora of options, which makes the result quite broad. Hence, a comprehensive description is imperative to elicit more precise and relevant outputs~\cite{yang2019exploring}.


\subsection{Be clear and precise}
The second basic prompt method is ``to be clear and precise". This involves formulating prompts that are unambiguous and specific, which can guide the model toward generating the desired output. 

Most LLM architectures are derived from an extensive array of textual data. It can be conceptualized as a combination of insights from a myriad of authors. When presented with a broad or undetailed prompt, its output predominantly exhibits a generic nature, which, while being applicable across a range of contexts, may not be optimal for any specific application. In contrast, a detailed and precise prompt enables the model to generate content that is more aligned with the unique requirements of the given scenario, as it reduces the model's uncertainty and guides it toward the correct response. 

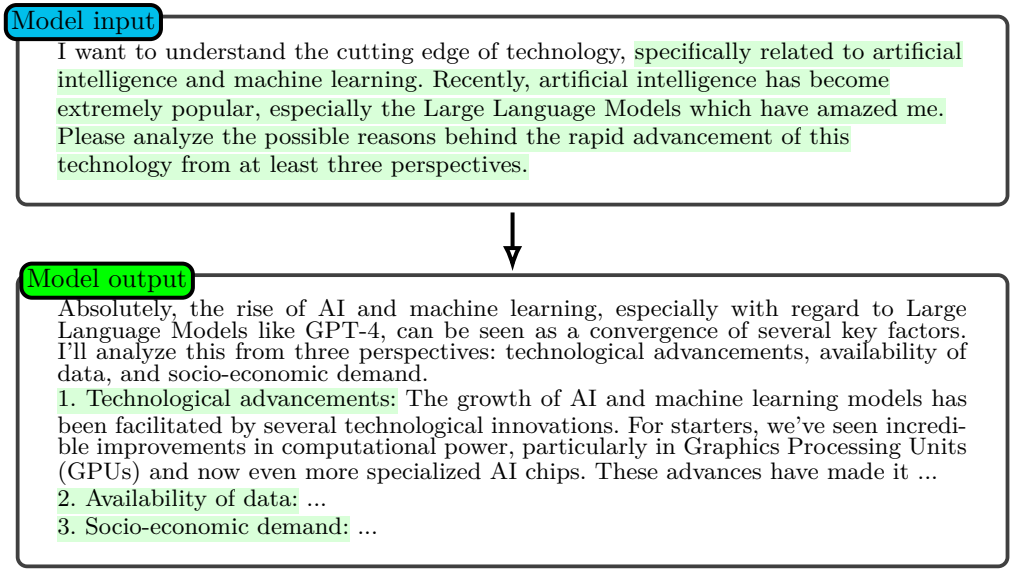
\begin{figure}[h]
\begin{spacing}{0.9}
\centering
\begin{minipage}{1\textwidth}
\large
\centering
\tikzmarknode{basic11}{}
\begin{tcolorbox}[colback=white]
\footnotesize
\raggedright
\textcolor{black}{\scriptsize I want to understand the cutting edge of technology, \hl{specifically related to artificial intelligence and machine learning. Recently, artificial intelligence has become extremely popular, especially the Large Language Models which have amazed me. Please analyze the possible reasons behind the rapid advancement of this technology from at least three perspectives.}}\tikzmark{a}
\end{tcolorbox}
\tikzmarknode{basic12}{}
\begin{tcolorbox}[colback=white]
\scriptsize
Absolutely, the rise of AI and machine learning, especially with regard to Large Language Models like GPT-4, can be seen as a convergence of several key factors. I'll analyze this from three perspectives: technological advancements, availability of data, and socio-economic demand.

\hl{1. Technological advancements:} The growth of AI and machine learning models has been facilitated by several technological innovations. For starters, we've seen incredible improvements in computational power, particularly in Graphics Processing Units (GPUs) and now even more specialized AI chips. These advances have made it ...

\hl{2. Availability of data:} ...

\hl{3. Socio-economic demand:} ...
\end{tcolorbox}
\end{minipage}\hfill
\begin{tikzpicture}[overlay,remember picture,-{Latex[open, scale=1]},shorten >=5pt,shorten <=5pt,line width=1.5pt]
\coordinate (start) at ([xshift=0pt, yshift=-79pt]basic11.south);
\coordinate (end) at ([xshift=0pt, yshift=-17pt]$(start)!0.9!(basic12.north)$); 
\draw[white, line width=0.6mm] (start) -- (end);
\draw[black, -{Latex[open, scale=1]}] (start) -- (end);
\node[anchor=north east, draw, fill=mylightblue, text=black, minimum width=1cm, minimum height=0.4cm, inner sep=2pt, rounded corners] at ([xshift=-4.6cm, yshift=-0.2cm]basic11) {Model input};
\node[anchor=north east, draw, fill=green, text=black, minimum width=1cm, minimum height=0.4cm, inner sep=2pt, rounded corners] at ([xshift=-4.18cm, yshift=-0.2cm]basic12) {Model output};

\end{tikzpicture}
\caption{\scriptsize A clearer and more precise prompt.}
\label{A clearer and more precise prompt.}
\end{spacing}
\end{figure}

For instance, as shown in Figure \ref{A clearer and more precise prompt.}, instead of asking a vague requirement such as ``I want to understand the cutting edge of technology.", a more precise prompt would be ``I want to understand the cutting edge of technology, specifically related to artificial intelligence and machine learning...".

\subsection{Role-prompting}
\begin{figure}[h]
\begin{spacing}{1.3}
\centering
\begin{minipage}{0.45\textwidth}
\large
\tikzmarknode{a1}{}
\begin{tcolorbox}[colback=white]
\footnotesize
\raggedright
\textcolor{black}{\scriptsize You are an expert in artificial intelligence specializing in Large Language Models...}\tikzmark{leftbox}
\end{tcolorbox}
\end{minipage}
\hfill
\begin{minipage}{0.45\textwidth}
\tikzmarknode{a2}{}
\begin{tcolorbox}[colback=white]
\scriptsize
Here are five basic methods to quickly get started with Large Language Models:

1. Experiment with web-based interfaces:

...
\tikzmark{rightbox}
\end{tcolorbox}
\end{minipage}

\begin{tikzpicture}[overlay,remember picture,-{Latex[open, scale=1]},shorten >=5pt,shorten <=5pt,line width=1.5pt]
\draw[black] ([xshift=2.7cm, yshift=0.45cm]pic cs:leftbox) -- ([xshift=-0.80cm,yshift=1.1cm]pic cs:rightbox);
\node[anchor=north east, draw, fill=mylightblue, text=black, minimum width=1cm, minimum height=0.4cm, inner sep=2pt, rounded corners] at ([xshift=2.2cm, yshift=-0.2cm]a1) {A role prompt};
\node[anchor=north west, draw, fill=green, text=black, minimum width=1cm, minimum height=0.4cm, inner sep=2pt, rounded corners] at ([xshift=-0.3cm, yshift=-0.2cm]a2) {Model output};
\end{tikzpicture}

\caption{{\scriptsize Role prompting example.}}
\label{Role prompting example.}
\end{spacing}
\end{figure}
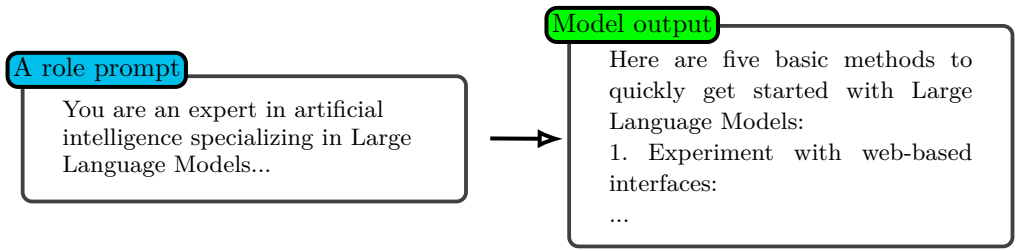

Role-prompting is another fundamental method in prompt engineering. It involves giving the model a specific role to play, such as a helpful assistant or a knowledgeable expert \cite{VISAR2023}. This method can be particularly effective in guiding the model's responses and ensuring that they align with the desired output. For instance, if the model is prompted to act as a historian, it is more likely to provide a detailed and contextually accurate response when asked about a historical event \cite{GuidedScenarios2023}. Another brief example is shown in Figure \ref{Role prompting example.}.

\subsection{Use of triple quotes to separate}
In prompt engineering, the use of triple quotes is a technique used to separate different parts of a prompt or to encapsulate multi-line strings. This technique is particularly useful when dealing with complex prompts that include multiple components or when the prompt itself contains quotes, which makes the model understand one's instructions better \cite{triplequote2023}.

\subsection{Try several times}
Due to the non-deterministic nature of LLMs, it is often beneficial to try several times when generating responses. This technique, often referred to as ``resampling", involves running the model multiple times with the same prompt and selecting the best output. This approach can help overcome the inherent variability in the model's responses and increase the chances of obtaining a high-quality output \cite{holtzman2020curious}.

\subsection{One-shot or few-shot prompting\label{subsec:one-shot-few-shot}}

One-shot and few-shot prompting are two important techniques in prompt engineering. One-shot prompting refers to the method where the model is given a single example to learn from, while few-shot \cite{logan-iv-etal-2022-cutting} prompting provides the model with multiple examples \cite{shyr2023identifying}. The choice between one-shot and few-shot prompting often depends on the complexity of the task and the capability of the model. For instance, for simple tasks or highly capable models, one-shot prompting might be sufficient. An example is shown in Figure \ref{Comparison of Standard prompt and One-shot prompt.}. However, for more complex tasks or less capable models, few-shot prompting can provide additional context and guidance, thereby improving the model's performance.

\definecolor{mylightblue}{RGB}{0,191,235} 
\definecolor{anotherlightblue}{RGB}{204,229,255} 
\sethlcolor{anotherlightblue}
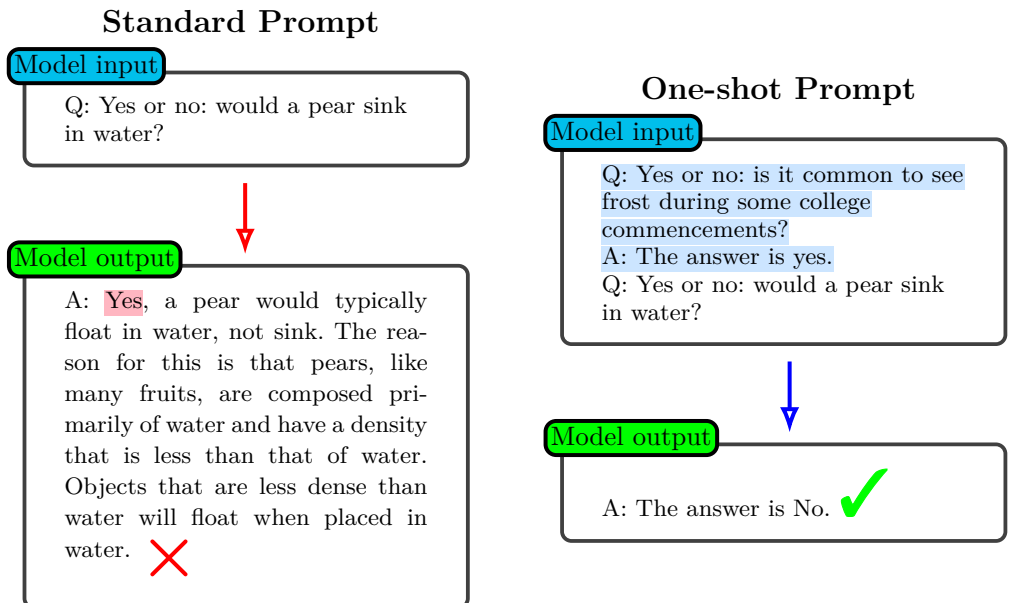
\begin{figure}[h]
\begin{spacing}{1.3}
\centering
\begin{minipage}{.45\textwidth}
\large
\centering
\textbf{Standard Prompt}
\tikzmarknode{a1}{}
\begin{tcolorbox}[colback=white]
\footnotesize
\raggedright
\textcolor{black}{\scriptsize Q: Yes or no: would a pear sink in water?}\tikzmark{a}
\end{tcolorbox}
\tikzmarknode{a2}{}
\definecolor{LightPink}{RGB}{255,182,193} 
\sethlcolor{LightPink}
\begin{tcolorbox}[colback=white]
\scriptsize
A: \hl{Yes}, a pear would typically float in water, not sink. The reason for this is that pears, like many fruits, are composed primarily of water and have a density that is less than that of water. Objects that are less dense than water will float when placed in \raisebox{1.67ex}{water.}
{\Huge \textcolor{red}{\( \boldsymbol{\times} \)}}
\end{tcolorbox}
\end{minipage}\hfill
\begin{minipage}{.45\textwidth}
\large
\centering
\textbf{One-shot Prompt}
\tikzmarknode{b1}{}
\begin{tcolorbox}[colback=white]
\footnotesize
\raggedright
\textcolor{black}{\scriptsize \hl{Q: Yes or no: is it common to see frost during some college commencements?\\
A: The answer is yes.}\\
Q: Yes or no: would a pear sink in water?}\tikzmark{b}
\end{tcolorbox}
\tikzmarknode{b2}{}
\begin{tcolorbox}[colback=white]
\scriptsize
\raggedright
A: The answer is No.
{\Huge \textcolor{green}{\pdfrender{StrokeColor=green,TextRenderingMode=2}{\ding{51}}}}
\end{tcolorbox}
\end{minipage}
\caption{\scriptsize Comparison of standard prompt and one-shot prompt.}
\label{Comparison of Standard prompt and One-shot prompt.}
\begin{tikzpicture}[overlay,remember picture,-{Latex[open, scale=1]},shorten >=5pt,shorten <=5pt,line width=1.5pt]
\draw[red] ([yshift=22.5ex,xshift=-10.7cm]pic cs:a) -- ++(0,-1.3cm); 
\draw[blue] ([yshift=-2.5ex,xshift=1.15cm]pic cs:b) -- ++(0,-1.3cm);
\node[anchor=north east, draw, fill=mylightblue, text=black, minimum width=1cm, minimum height=0.4cm, inner sep=2pt, rounded corners] at ([xshift=-2.95cm, yshift=-0.2cm]a1) {Model input};
\node[anchor=north east, draw, fill=green, text=black, minimum width=1cm, minimum height=0.4cm, inner sep=2pt, rounded corners] at ([xshift=-0.7cm, yshift=-0.2cm]a2) {Model output};
\node[anchor=north east, draw, fill=mylightblue, text=black, minimum width=1cm, minimum height=0.4cm, inner sep=2pt, rounded corners] at ([xshift=-2.95cm, yshift=-0.2cm]b1) {Model input};
\node[anchor=north east, draw, fill=green, text=black, minimum width=1cm, minimum height=0.4cm, inner sep=2pt, rounded corners] at ([xshift=-0.85cm, yshift=-0.2cm]b2) {Model output};
\end{tikzpicture}
\end{spacing}
\end{figure}

However, as stated in \cite{Reynolds2021}, ``examples don’t always help", meaning that zero-shot prompting may have better output in some scenarios. Zero-shot prompting \cite{10.1145/3560815, kojima2022large}, in the context of prompt-based learning, involves using a pre-trained LLM to perform tasks without any specific training for those tasks. The model relies on its general knowledge, acquired during pre-training, to generate predictions based on cleverly crafted prompts. This allows the LLMs to handle new tasks with no additional task-specific data, making it adaptable to scenarios with minimal labeled data. \cite{Reynolds2021} investigated the intricacies of how large generative language models, such as GPT-3, respond to prompts. One of the significant findings from this paper is that zero-shot prompts can, in certain scenarios, outperform few-shot prompts. This suggests that the role of few-shot examples might not be as much about teaching the model a new task (meta-learning) but rather guiding it to recall a task it has already learned. This insight is crucial as it challenges the conventional wisdom that more examples always lead to better performance \cite{brownGPT3}. In the context of one-shot or few-shot prompting, it is essential to understand that while examples can guide the model, they do not always enhance its performance. Sometimes, a well-crafted zero-shot prompt can be more effective than providing multiple examples \cite{liu2020multi}.

\subsection{LLMs settings: temperature and top-$p$}
The settings of LLMs, such as the temperature and top-$p$, play a crucial role in the generation of responses. The temperature parameter controls the randomness of the generated output: a lower temperature leads to more deterministic outputs \cite{ACKLEY1985147, ficler2017controlling}. The top-$p$ parameter, on the other hand, controls the nucleus sampling \cite{holtzman2020curious}, which is a method to add randomness to the model's output \cite{riga2023semeval}. Adjusting these parameters can significantly affect the quality and diversity of the model's responses, making them essential tools in prompt engineering. However, it has been noted that certain models, exemplified by ChatGPT, do not permit the configuration of these hyperparameters, barring instances where the Application Programming Interface (API) is employed. \cite{liesenfeld2024rethinking} ranks several AI text generators and text-to-image systems in terms of various openness metrics, including the accessibility of their API and model parameters.

\section{Advanced methodologies\label{sec:advanced_methodologies}}
The foundational methods from the previous section can help us produce satisfactory outputs. However, experiments indicate that when using LLMs for complex tasks such as analysis or reasoning, the accuracy of the model's outputs still has room for improvement. In this section, advanced techniques of prompt engineering will be introduced to guide the model in generating more specific, accurate, and high-quality content.

\subsection{Chain-of-thought}
\label{sec3.1}
The concept of ``Chain-of-Thought" (CoT) prompting \cite{wei2022chain} in LLMs is a relatively new development, which has been shown to significantly improve the accuracy of LLMs on various logical reasoning tasks \cite{wu2023analyzing, zhang2023automatic, lewkowycz2022solving}. CoT prompting involves providing intermediate reasoning steps to guide the model's responses, which can be facilitated through simple prompts such as ``Let's think step by step" or through a series of manual demonstrations, each composed of a question and a reasoning chain that leads to an answer \cite{zhou2022teaching, lee2023teaching}. It also provides a clear structure for the model's reasoning process, making it easier for users to understand how the model arrived at its conclusions.

\cite{wei2022large} illustrates the application of CoT prompting to medical reasoning, showing that it can effectively elicit valid intermediate reasoning steps from LLMs. \cite{zhang2023chain} introduces the concept of Self-Education via Chain-of-Thought Reasoning (SECToR), and argues that, in the spirit of reinforcement learning, LLMs can successfully teach themselves new skills by chain-of-thought reasoning. In another study, \cite{cobbe2021training} used CoT prompting to train verifiers to solve math word problems, demonstrating the technique's potential in educational applications. \cite{huang2023language} proposed a multimodal version of CoT, called Multimodal-CoT, to handle more complex, multimodal tasks beyond simple text-based tasks, such as visual tasks, further expanding the potential applications of CoT. Furthermore, many works are building upon the CoT framework, for instance, \cite{shum-etal-2023-automatic} introduces Automate-CoT, an automated approach for augmenting and selecting rationale chains to enhance the reasoning capabilities of LLMs, thereby reducing the reliance on manually crafted CoT prompts.

\subsubsection{Zero-shot chain-of-thought}
The concept of ``Zero-Shot chain-of-thought" (Zero-shot-CoT) prompting is an advanced iteration of the CoT prompting mechanism, where the ``zero-shot" aspect implies that the model is capable of performing some reasoning without having seen any examples of the task during training. 

\cite{kojima2022large} discovered that the augmentation of queries with the phrase ``Let's think step by step" facilitated the generation of a sequential reasoning chain by LLMs. This reasoning chain subsequently proved instrumental in deriving more precise answers. This technique is based on the idea that the model, much like a human, can benefit from having more detailed and logical steps to process the prompt and generate a response.

\begin{figure}[h]
\begin{spacing}{1.0}
\centering
\begin{minipage}{1\textwidth}
\large
\centering
\tikzmarknode{basic11}{}
\begin{tcolorbox}[colback=white]
\footnotesize
\raggedright
\textcolor{black}{\scriptsize Imagine an infinitely wide entrance, which is more likely to pass through it, a military tank or a car? 
\hl{Let's think step by step.}}   
\end{tcolorbox}
\tikzmarknode{basic12}{}

\begin{tcolorbox}[colback=white]
\scriptsize
Given an infinitely wide entrance, both a military tank and a car could pass through without any issues related to the width. However, there are still several factors that could influence which is more likely to pass through:

\hl{1. Ground surface:} tanks are designed to handle rough terrains that cars may struggle with. If the ground leading to and from the entrance is too rugged or steep for a car, the tank would be more likely to pass through.

\hl{2. Weight restrictions:} if the entrance has a load limit due to the materials used in its construction, the car, being lighter, could be more likely to pass through. The average car weighs around 1.5 tons, while a military tank can weigh anywhere from 60 to 70 tons.

\hl{3. Height clearance:} if the entrance is infinitely wide but not infinitely tall, the height of the vehicle matters. The average height of a car is around 1.5 meters, while tanks can be around 2.4 meters. If the entrance is shorter than around 2.4 meters, a car would be more likely to pass through.

\hl{4. Regulations and permissions:}\\
...

In conclusion, without specific information about the conditions at the entrance, \hl{we would assume that a car is more likely to pass through}, simply because cars are more common, more likely to meet potential height and weight restrictions and face fewer regulatory barriers.
\end{tcolorbox}
\end{minipage}\hfill

\begin{tikzpicture}[overlay,remember picture,-{Latex[open, scale=0.5]},shorten >=5pt,shorten <=5pt,line width=1.5pt]
\coordinate (start) at ([xshift=0pt, yshift=-46pt]basic11.south);
\coordinate (end) at ([xshift=0pt, yshift=-17pt]$(start)!0.9!(basic12.north)$); 
\draw[white, line width=0.8mm] (start) -- (end);
\draw[black, -{Latex[open, scale=1]}] (start) -- (end);

\node[anchor=north east, draw, fill=mylightblue, text=black, minimum width=1cm, minimum height=0.4cm, inner sep=2pt, rounded corners] at ([xshift=-4.6cm, yshift=-0.2cm]basic11) {Model input};
\node[anchor=north east, draw, fill=green, text=black, minimum width=1cm, minimum height=0.4cm, inner sep=2pt, rounded corners] at ([xshift=-4.2cm, yshift=-0.2cm]basic12) {Model output};
\end{tikzpicture}
\caption{\scriptsize Adding ``Let's think step by step".}
\label{Add ``Let's think step by step".}
\end{spacing}
\end{figure}
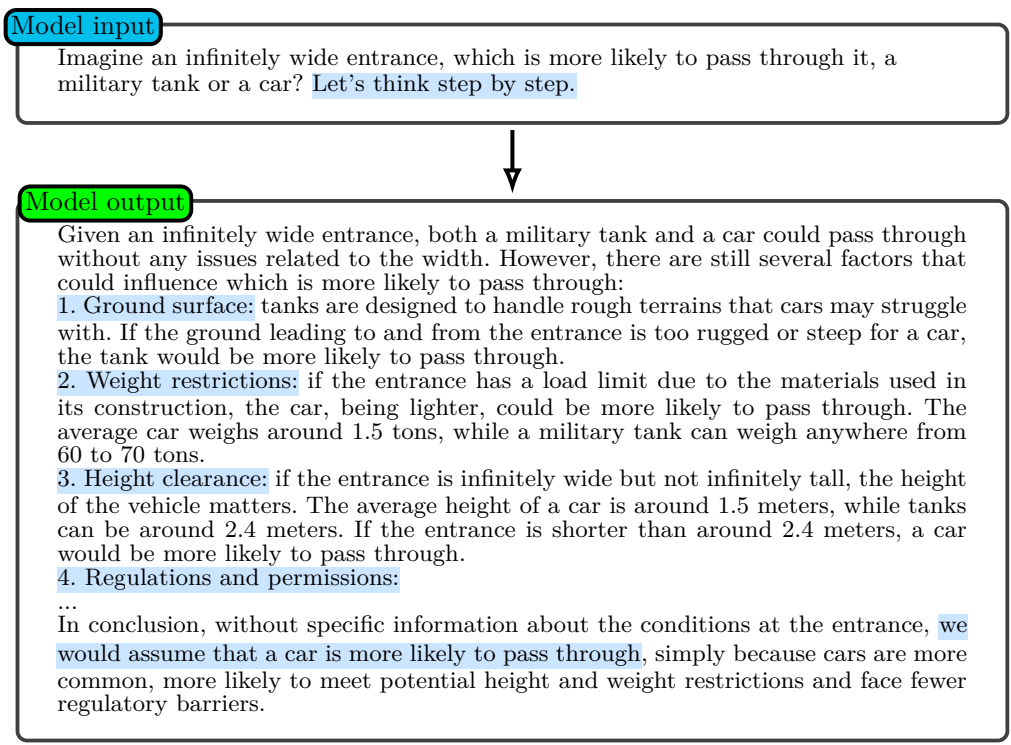

For instance, the standard prompt is illustrated in Figure \ref{Standard Prompt.}, while the appended phrase, ``Let's think step by step", is depicted in Figure \ref{Add ``Let's think step by step".}. Observations indicate that the incorporation of ``Let's think step by step" enhances the logical coherence and comprehensiveness of the model's response.

\begin{figure}[h]
\begin{spacing}{1.1}
\centering
\begin{minipage}{1\textwidth}
\large
\centering
\tikzmarknode{basic11}{}
\begin{tcolorbox}[colback=white]
\footnotesize
\raggedright
\textcolor{black}{\scriptsize Imagine an infinitely wide entrance, which is more likely to pass through it, a military tank or a car?}\tikzmark{a}
\end{tcolorbox}
\tikzmarknode{basic12}{}
\begin{tcolorbox}[colback=white]
\scriptsize
Given an ``infinitely wide entrance", both a military tank and a car can pass through it without any constraints due to width. Therefore, neither is more likely than the other based on width alone. Both can pass through with equal likelihood. 

However, if other factors such as ground conditions, accessibility, purpose, or authorization come into play, the likelihood might change. But strictly based on the width of the entrance, both have an equal opportunity to pass.
\end{tcolorbox}
\end{minipage}\hfill

\begin{tikzpicture}[overlay,remember picture,-{Latex[open, scale=1]},shorten >=5pt,shorten <=5pt,line width=1.5pt]
\coordinate (start) at ([yshift=-45pt]basic11.south);
\coordinate (end) at ([yshift=-17pt]$(start)!0.9!(basic12.north)$); 
\draw[white, line width=0.8mm] (start) -- (end);
\draw[black, -{Latex[open, scale=1]}] (start) -- (end);

\node[anchor=north east, draw, fill=mylightblue, text=black, minimum width=1cm, minimum height=0.4cm, inner sep=2pt, rounded corners] at ([xshift=-4.6cm, yshift=-0.2cm]basic11) {Model input};
\node[anchor=north east, draw, fill=green, text=black, minimum width=1cm, minimum height=0.4cm, inner sep=2pt, rounded corners] at ([xshift=-4.3cm, yshift=-0.2cm]basic12) {Model output};
\end{tikzpicture}

\caption{\scriptsize Standard prompt.}
\label{Standard Prompt.}
\end{spacing}
\end{figure}
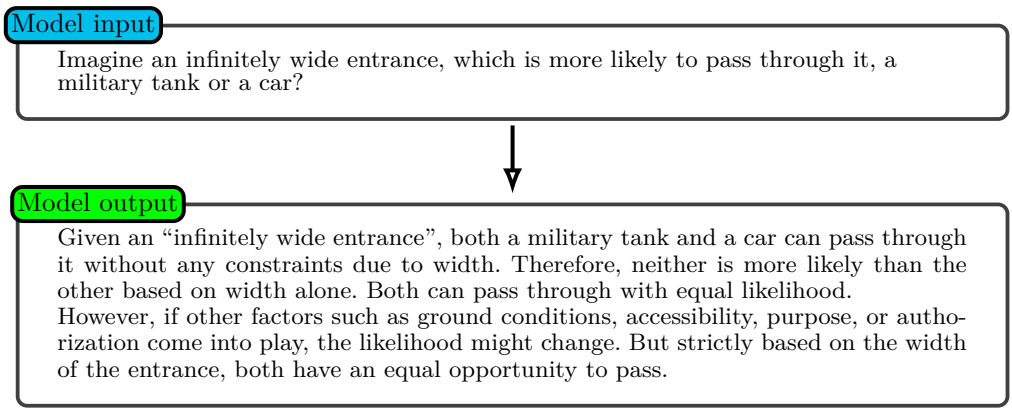

\subsubsection{Golden chain-of-thought}
\cite{del2023true} introduced the ``golden chain-of-thought", providing an innovative approach to generating responses to instruction-based queries. This methodology leverages a set of ``ground-truth chain-of-thought" solutions incorporated within the prompt, considerably simplifying the task for the model as it circumvents the necessity for independent CoT generation. Concurrently, a novel benchmark comprising detective puzzles has been designed, to assess the abductive reasoning capacities of LLMs, which is also considered an evaluation of the golden CoT. Finally, according to the experiment by \cite{del2023true}, in the context of the golden CoT, GPT-4 exhibits commendable performance, boasting an 83\% solve rate of puzzles in contrast to the 38\% solve rate of the standard CoT. 

However, the characteristics of the Golden CoT requiring the ``ground-truth chain-of-thought solutions" as an integral part of the prompt also signifies that the Golden CoT's contribution to solving such problems is limited, despite its high solve rate of 83\%.

\subsection{Self-consistency}
In the assessment of \textit{Instruct}GPT \cite{ouyang2022training} and GPT-3 \cite{brownGPT3} on a new synthetic QA dataset called Proof and Ontology-Generated Question-Answering (PrOntoQA) \cite{saparov2023language, tafjord-etal-2021-proofwriter}, it was observed that although the most extensive model exhibited capability in reasoning tasks, it encountered challenges in proof planning and the selection of the appropriate proof step amidst multiple options, which caused accuracy uncertainties \cite{saparov2023language}. Self-consistency is one of the methods for LLMs to solve this situation, which is an advanced prompting technique that aims to ensure the model's responses are consistent with each other \cite{wang2023selfconsistency, wei2022chain}. This method greatly increases the odds of obtaining highly accurate results. The principle of self-consistency in language models posits that for a complex reasoning problem, there can be multiple reasoning paths leading to the correct answer. In this approach, a language model generates a diverse set of reasoning paths for the same problem. The most accurate and consistent answer is then determined by evaluating and marginalizing across these varied paths, ensuring that the final answer reflects the convergence of multiple lines of thought.

The self-consistency method contains three steps. Firstly, prompt a language model using CoT prompting, then replace the “greedy decode” (1-Best) \cite{Radford2019LanguageMA,chowdhery2022palm} in CoT prompting by sampling from the language model’s decoder to generate a diverse set of reasoning paths, and finally, marginalize out the reasoning paths and aggregate by choosing the most consistent answer in the final answer set.

It is noteworthy that self-consistency can be harmoniously integrated with most sampling algorithms, including but not limited to, temperature sampling \cite{ACKLEY1985147, ficler2017controlling}, \mbox{top-$k$} sampling \cite{fan2018hierarchical, holtzman2018learning, Radford2019LanguageMA}, and nucleus sampling \cite{holtzman2020curious}. Nevertheless, such an operation may necessitate the invocation of the model's API to fine-tune these hyperparameters. In light of this, an alternative approach could be to allow the model to generate results employing diverse reasoning paths, and then generate a diverse set of candidate reasoning paths. The response demonstrating the highest degree of consistency across the various reasoning trajectories is then more inclined to represent the accurate solution \cite{huang2022large}.

\cite{shum2023automatic,bender2021dangers} have shown that self-consistency enhances outcomes in arithmetic, commonsense, and symbolic reasoning tasks. Furthermore, in practice, self-consistency can be combined with other techniques to further enhance the model's performance. \cite{khalifa2023discriminator} found that combining self-consistency with a discriminator-guided multi-step reasoning approach significantly improved the model's reasoning capabilities.

\subsection{Generated knowledge}
The ``generated knowledge" \cite{liu-etal-2022-generated} approach in prompt engineering is a technique that leverages the ability of LLMs to generate potentially useful information about a given question or prompt before generating a final response. This method is particularly effective in tasks that require commonsense reasoning, as it allows the model to generate and utilize additional context that may not be explicitly present in the initial prompt.

\begin{figure}[h]
\begin{spacing}{1.1}
\centering
\begin{minipage}{1\textwidth}
\large
\centering
\tikzmarknode{basic11}{}
\begin{tcolorbox}[colback=white]
\scriptsize
\raggedright
\textcolor{black}
{``}{``}{``}

\hl{Detailed size analysis: military tanks vs. cars}

\hl{1. Dimensions and weight:}

    \hl{- Military tanks: typically, main battle tanks (like the M1 Abrams or T-90) have widths ranging from 10 to 13 feet, lengths from 25 to 35 feet (including the main gun), and weights from 60 to 70 tons.}
    
    \hl{- Cars: most consumer cars have a width of about 6 to 8 feet, a length of 12 to 20 feet, and weigh between 1 to 3 tons. Some large SUVs and trucks can exceed these figures but are still substantially smaller and lighter than tanks.}
    
\hl{2. Maneuverability \& design...}

"""\\

Use the above information to answer the following question: 
Imagine an infinitely wide entrance, which is more likely to pass through it, a military tank or a car?\tikzmark{a}
\end{tcolorbox}
\tikzmarknode{basic12}{}

\begin{tcolorbox}[colback=white]
\scriptsize
Given the information provided about the dimensions, design, and the factors influencing whether an object can pass through an infinitely wide entrance, we can infer the following:

1. Height of the object: the height is not specified for either tanks or cars. However, given that tanks are built for rough terrains and often have a turret and other equipment on top, they generally have a taller profile compared to most cars. If the entrance had a restrictive height, cars would generally have an advantage here.

2. Structural integrity: ...

3. Ground clearance and maneuverability: \\
...

Conclusion:
\hl{If the entrance height is not restrictive and the surface is rough, a military tank would be more likely to pass through it without issues. On the other hand, if the entrance is low-height or has tight squeezes that could compromise the structural integrity of larger, robust vehicles, a car might be at an advantage.}

However, given no height constraints and only the factor of width (since it is ``infinitely wide"), both a military tank and a car would easily pass through it. In such a scenario, the distinguishing factor would be the terrain of the entrance. If it is rough or uneven, the tank has the advantage. If it is smooth, both would pass with relative ease.
\end{tcolorbox}
\end{minipage}\hfill

\begin{tikzpicture}[overlay,remember picture,-{Latex[open, scale=0.3]},shorten >=5pt,shorten <=5pt,line width=1.5pt]

\coordinate (start) at ([xshift=0pt, yshift=-165pt]basic11.south);
\coordinate (end) at ([xshift=0pt, yshift=-19pt]$(start)!0.9!(basic12.north)$); 
\draw[white, line width=0.8mm] (start) -- (end);
\draw[black, -{Latex[open, scale=1]}] (start) -- (end);

\node[anchor=north east, draw, fill=mylightblue, text=black, minimum width=1cm, minimum height=0.4cm, inner sep=2pt, rounded corners] at ([xshift=-4.6cm, yshift=-0.2cm]basic11) {Model input};
\node[anchor=north east, draw, fill=green, text=black, minimum width=1cm, minimum height=0.4cm, inner sep=2pt, rounded corners] at ([xshift=-4.4cm, yshift=-0.2cm]basic12) {Model output};
\end{tikzpicture}

\caption{{\scriptsize Combining the generated knowledge with the question (Step2).}}
\label{Combine the Generated Knowledge with the question(Step2).}
\end{spacing}
\end{figure}
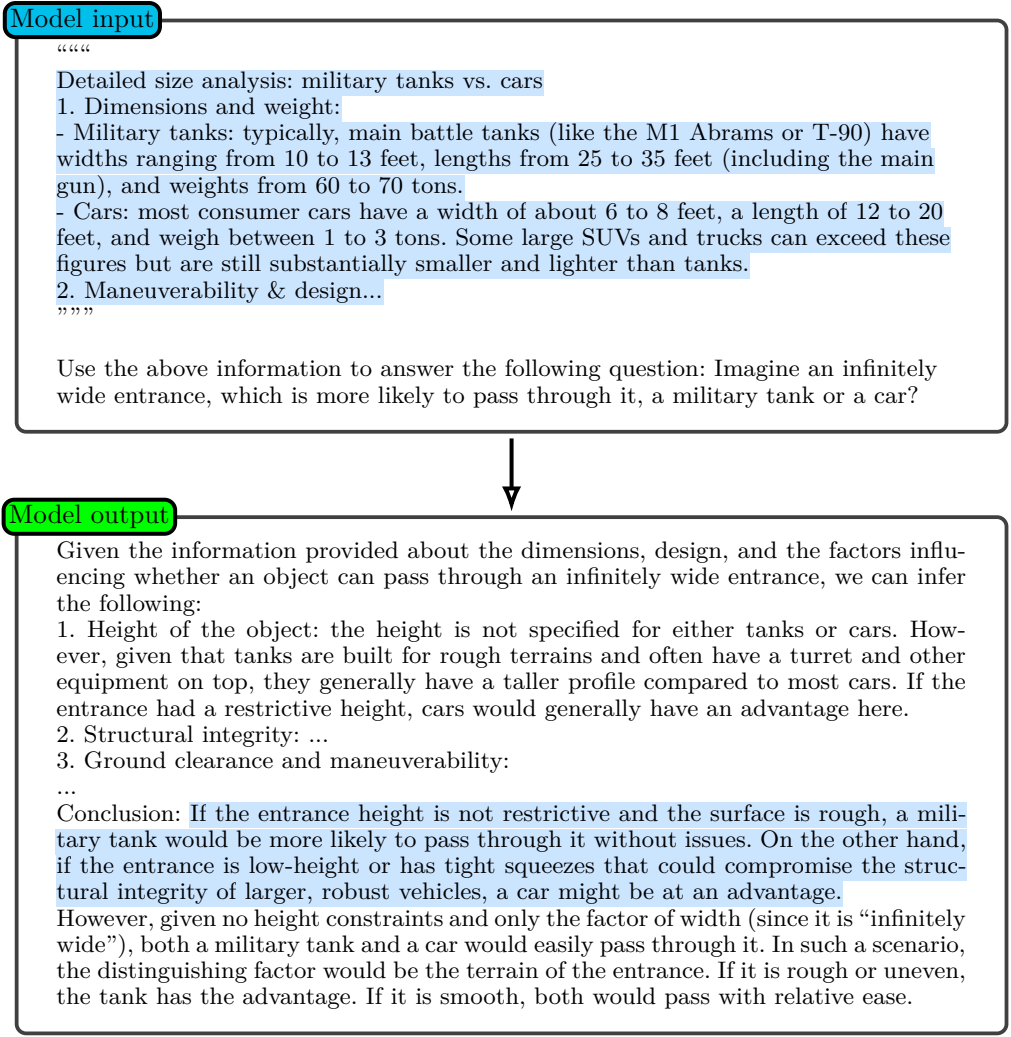

As exemplified in Figure \ref{Standard Prompt.}, when posing the query to the model, ``Imagine an infinitely wide entrance, which is more likely to pass through it, a military tank or a car?", standard prompts predominantly yield responses that neglect to factor in the ``entrance height". Conversely, as delineated in Figure \ref{Generate Knowledge(Step1).} and Figure \ref{Combine the Generated Knowledge with the question(Step2).}, prompting the model to first generate pertinent information and subsequently utilizing generated information in the query leads to outputs with augmented logical coherence and comprehensiveness. Notably, this approach stimulates the model to account for salient factors such as ``entrance height".

\begin{figure}[h]
\begin{spacing}{1.1}
\centering
\begin{minipage}{1\textwidth}
\large
\centering
\tikzmarknode{basic11}{}
\begin{tcolorbox}[colback=white]
\footnotesize
\raggedright
\textcolor{black}{\scriptsize Generate two key analyses related to detailed size data on military tanks and cars, and then generate three key influencing factors regarding whether an object can pass through an infinitely wide entrance.}\tikzmark{a}
\end{tcolorbox}
\tikzmarknode{basic12}{}
\begin{tcolorbox}[colback=white]
\scriptsize
\hl{Detailed size analysis: military tanks vs. cars}

\hl{1. Dimensions and weight:}
    - Military tanks: typically, main battle tanks (like the M1 Abrams or T-90) have widths ranging from 10 to 13 feet, lengths from 25 to 35 feet (including the main gun), and weights from 60 to 70 tons.
    - Cars: most consumer cars have a width of about 6 to 8 feet, a length of 12 to 20 feet, and weigh between 1 to 3 tons. Some large SUVs and trucks can exceed these figures but are still substantially smaller and lighter than tanks.

\hl{2. Maneuverability \& design:}
    - Military tanks: tanks are designed to traverse rough terrains and have tracks that distribute weight and prevent the vehicle from getting easily stuck in muddy or uneven ground. However, their sheer weight and size limit their maneuverability in urban or congested settings.
    - Cars: designed for paved roads, cars are generally more agile and can navigate tight spaces more efficiently than tanks. Their design emphasizes aerodynamics and comfort, unlike tanks that prioritize armor and firepower.\\
    
\hl{Three key influencing factors on whether an object can pass through an infinitely wide entrance:}

\hl{1. Height of the object:} the height is critical. An infinitely wide entrance does not give information about its height. If the object is taller than the entrance, it won't pass through.

\hl{2. Structural integrity:} depending on the material and design of an object, squeezing or forcing it through an entrance could compromise its structural integrity. For instance, certain tanks have reactive armor or sensitive equipment that may be damaged if scraped or bumped forcefully.

\hl{3. Ground clearance and maneuverability:} ...

\end{tcolorbox}
\end{minipage}\hfill

\begin{tikzpicture}[overlay,remember picture,-{Latex[open, scale=0.7]},shorten >=5pt,shorten <=5pt,line width=1.5pt]
\coordinate (start) at ([yshift=-54pt]basic11.south);
\coordinate (end) at ([yshift=-20pt]$(start)!0.9!(basic12.north)$); 
\draw[white, line width=0.8mm] (start) -- (end);
\draw[black, -{Latex[open, scale=1]}] (start) -- (end);

\node[anchor=north east, draw, fill=mylightblue, text=black, minimum width=1cm, minimum height=0.4cm, inner sep=2pt, rounded corners] at ([xshift=-4.6cm, yshift=-0.2cm]basic11) {Model input};
\node[anchor=north east, draw, fill=green, text=black, minimum width=1cm, minimum height=0.4cm, inner sep=2pt, rounded corners] at ([xshift=-4.35cm, yshift=-0.2cm]basic12) {Model output};
\end{tikzpicture}

\caption{\scriptsize Generating knowledge (Step1).}
\label{Generate Knowledge(Step1).}
\end{spacing}
\end{figure}
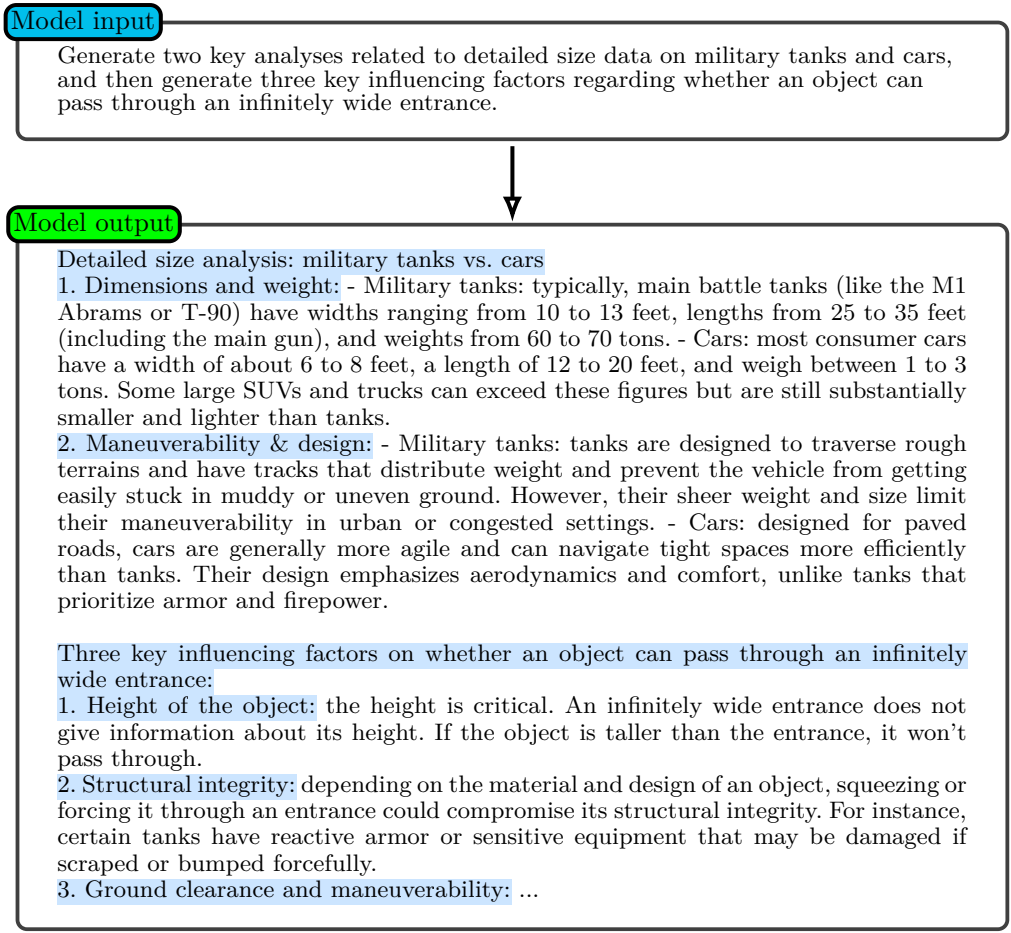

\subsection{Least-to-most prompting}

\begin{figure}[h]
  \centering
  \includegraphics[width=\linewidth]{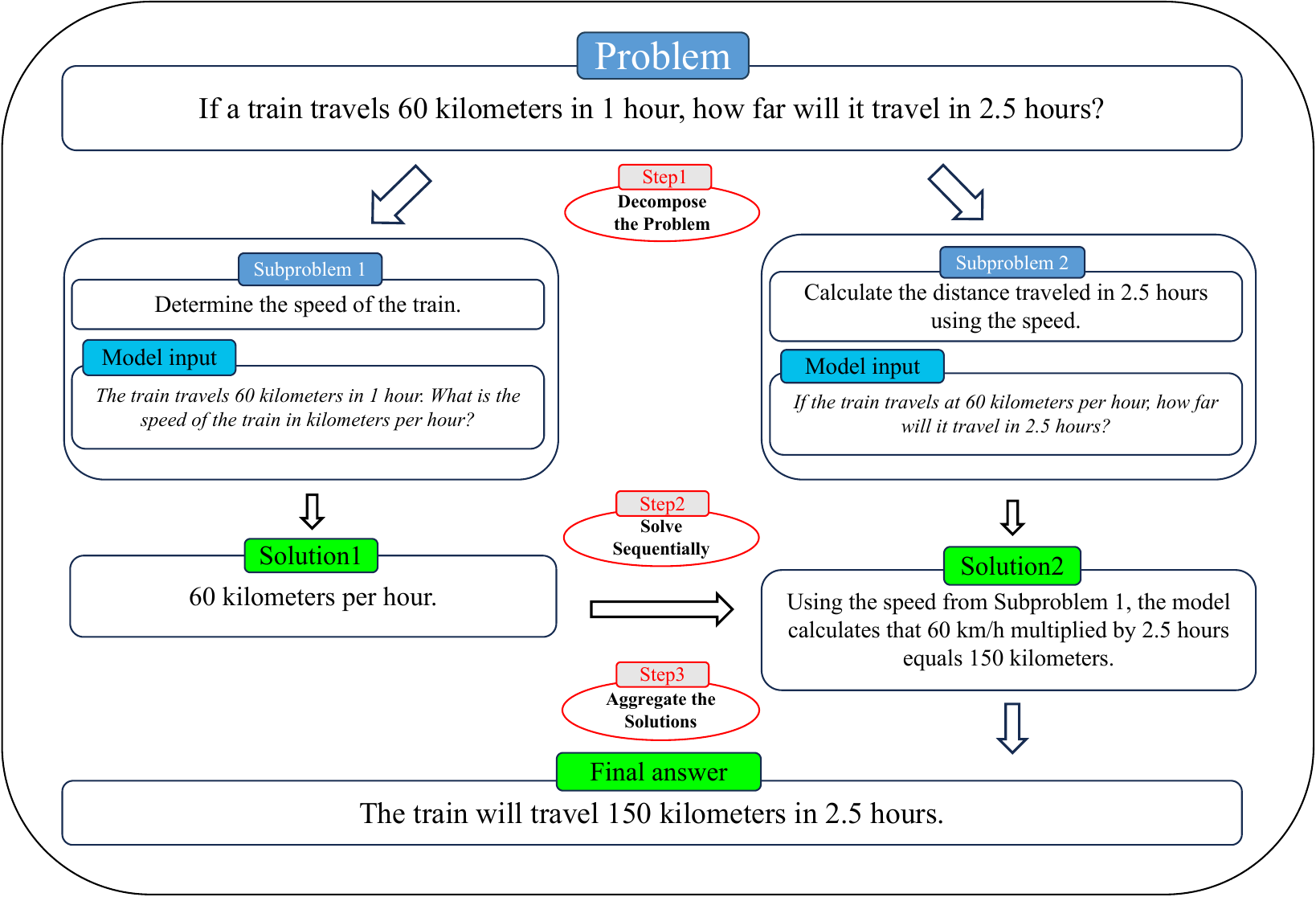}
  \caption{Illustration of Least-to-Most Prompting Applied to a Mathematical Word.}
  \label{Least-to-most prompting}
\end{figure}

The concept of ``least-to-most prompting" \cite{zhou2023leasttomost} is an advanced method that involves decomposing a complex problem into a series of simpler subproblems, which are then addressed sequentially. The foundational premise of this approach is to systematically break down intricate tasks into manageable components. Each subproblem is solved in turn, with the solution to each one serving as a building block for the next. This step-by-step process not only simplifies the problem-solving task for the language model but also enables it to handle more complex challenges than those typically managed by traditional prompting techniques. Figure \ref{Least-to-most prompting} is an illustration of least-to-most prompting applied to a mathematical word. This figure illustrates the least-to-most prompting approach by systematically breaking down a complex problem into a series of simpler subproblems. The process begins with the decomposition of the main problem—calculating the distance a train travels in 2.5 hours—into two sequential subproblems. First, the model is prompted to determine the train's speed, and then it uses this information to calculate the distance traveled. Each subproblem is solved in sequence, with the solution to the first subproblem feeding into the second. The solutions are then aggregated to arrive at the final answer. This method emphasizes the key principles of problem decomposition and sequential problem solving, enabling the model to manage and solve complex tasks more effectively.

Upon rigorous experimentation in domains including symbolic manipulation, compositional generalization, and mathematical reasoning, \cite{zhou2023leasttomost} substantiate that the least-to-most prompting paradigm exhibits the capacity to generalize across challenges of greater complexity than those initially presented in the prompts. \cite{gao2023pal} introduced Program Aided Language models (PAL), using the LLMs to read natural language problems and generate programs as the intermediate reasoning steps. By using least to most prompting, PAL shows enhancement on GSM8K \cite{cobbe2021training} and SVAMP \cite{patel2021nlp}, which are benchmarks about complex mathematical problems for LLMs.

\subsection{Tree of thoughts}

The ``tree of thoughts" (ToT) prompting technique in LLMs is an advanced method that employs a structured approach to guide LLMs in their reasoning and response generation processes. It enhances problem-solving by exploring multiple reasoning paths, termed `thoughts'. Unlike traditional linear prompts, ToT allows LLMs to consider various possible solutions and strategies, including looking ahead, backtracking, and self-evaluation, making it more interactive and adaptable to the complexity of the task at hand. This approach fosters more dynamic and deliberate decision-making in complex problem-solving tasks, moving beyond a rigid, hierarchical prompt structure to a more flexible and adaptive reasoning process \cite{yao2023tree}.
The ToT approach, for instance, when applied to complex mathematical problem-solving, prompts the model to generate various potential solutions and evaluate them, rather than simply asking for a solution.

\cite{yao2023tree} demonstrates that this formulation is more versatile and can handle challenging tasks where standard prompts might fall short. Another research by \cite{long2023large} further emphasizes the potential of this technique in enhancing the performance of LLMs by structuring their thought processes.

\cite{hulbert2023bard} introduces the ``tree-of-thought prompting", an approach that assimilates the foundational principles of the ToT frameworks and transforms them into a streamlined prompting methodology. This technique enables LLMs to assess intermediate cognitive constructs within a singular prompt. An exemplar ToT prompt is delineated in Figure~\ref{A sample ToT prompt.}.

\begin{figure}[h]
\begin{spacing}{1.1}
\FloatBarrier
\centering
\begin{minipage}{1\textwidth}
\large
\centering
\tikzmarknode{basic11}{}
\begin{tcolorbox}[colback=white]
\scriptsize
Imagine three different experts answering this question.\\
All experts will write down 1 step of their thinking,\\
then share it with the group.\\
Then all experts will go on to the next step, etc.\\
If any expert realizes they're wrong at any point then they leave.\\
The question is...
\end{tcolorbox}
\end{minipage}\hfill
\begin{tikzpicture}[overlay,remember picture,-{Latex[open, scale=1]},shorten >=5pt,shorten <=5pt,line width=2pt]
\node[anchor=north east, draw, fill=mylightblue, text=black, minimum width=1cm, minimum height=0.4cm, inner sep=2pt, rounded corners] at ([xshift=-2.45cm, yshift=-0.2cm]basic11) {Tree of thoughts prompting};
\end{tikzpicture}
\caption{\scriptsize A sample ToT prompt \cite{hulbert2023bard}.}
\label{A sample ToT prompt.}
\end{spacing}
\end{figure}

\subsection{Graph of thoughts}
Unlike the ``chain-of-thoughts" or ``tree of thoughts" paradigms, the ``graph of thoughts" (GoT) framework \cite{besta2023graph} offers a more intricate method of representing the information generated by LLMs. The core concept behind GoT is to model this information as an arbitrary graph. In this graph, individual units of information, termed ``LLM thoughts", are represented as vertices. The edges of the graph, on the other hand, depict the dependencies between these vertices. This unique representation allows for the combination of arbitrary LLM thoughts, thereby creating a synergistic effect in the model's outputs.

In the context of addressing intricate challenges, LLMs utilizing the GoT framework might initially produce several autonomous thoughts or solutions. These individual insights can subsequently be interlinked based on their pertinence and interdependencies, culminating in a detailed graph. This constructed graph permits diverse traversal methods, ensuring the final solution is both precise and comprehensive, encompassing various dimensions of the challenge.

The efficacy of the GoT framework is anchored in its adaptability and the profound insights it can yield, particularly for intricate issues necessitating multifaceted resolutions. Nonetheless, it is imperative to recognize that while GoT facilitates a systematic approach to problem-solving, it also necessitates a profound comprehension of the subject matter and meticulous prompt design to realize optimal outcomes \cite{wang2023llm}.

\subsection{Decomposed prompting}
Decomposed Prompting (DECOMP) \cite{khot2023decomposedpromptingmodularapproach} is a modular approach designed to tackle complex tasks by breaking them down into simpler, manageable sub-tasks. This methodology leverages the capabilities of LLMs by creating a systematic process where each sub-task is handled by specialized handlers. The approach not only simplifies the problem-solving process but also enhances the flexibility and efficiency of task handling.

\begin{figure}[h]
  \centering
  \includegraphics[width=\linewidth]{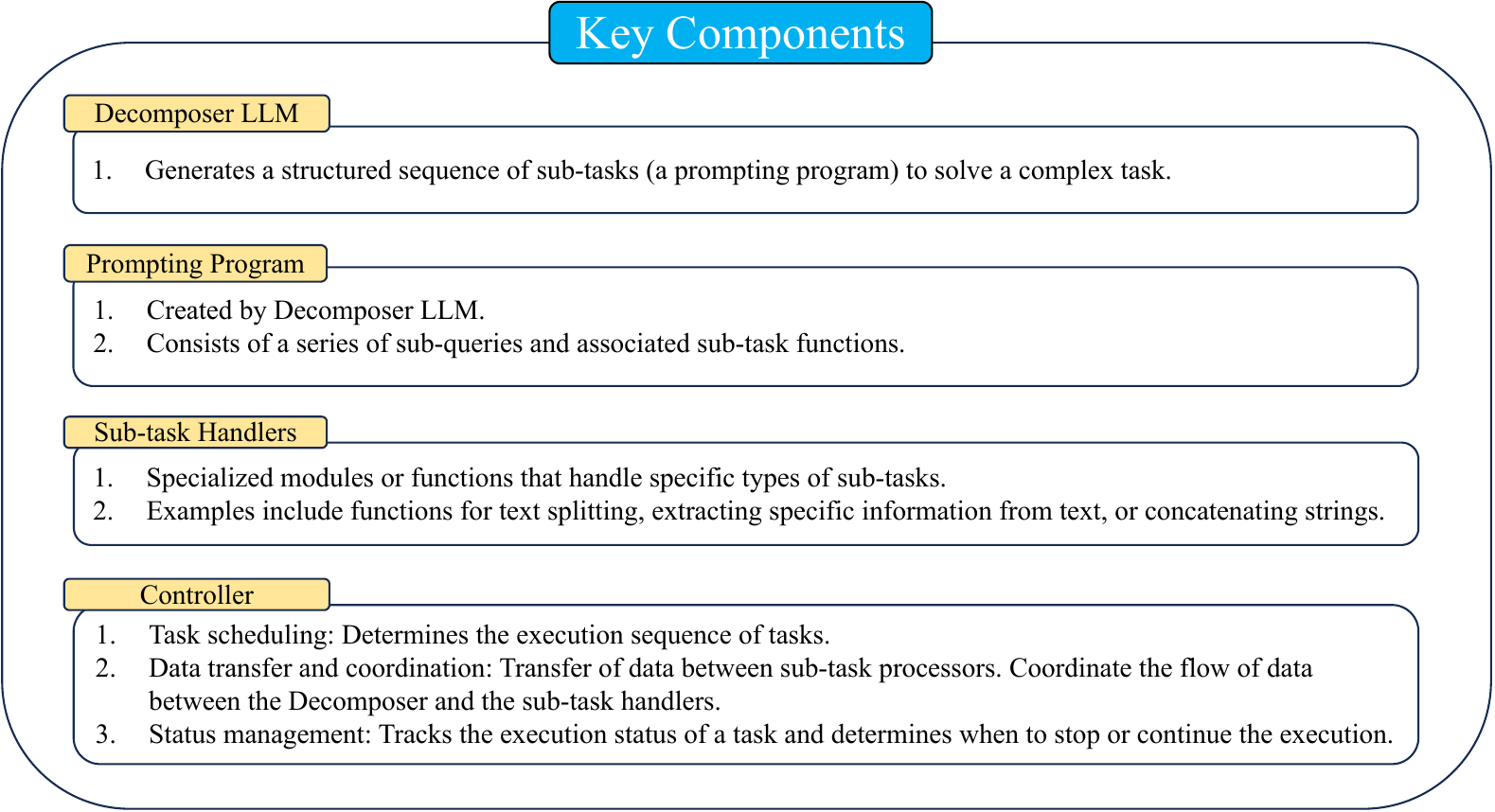}
  \caption{Key components of DECOMP}
  \label{key_component}
\end{figure}

Four key components of this method are shown in Figure \ref{key_component}. The core of DECOMP involves a decomposer LLM that generates a prompting program $P$ for a complex task $Q$. The program $P$ is a sequence of steps, each step directing a simpler sub-query to a function within an auxiliary set of sub-task functions $F$. The program can be represented as:
\[
P = \{ (f_1, Q_1, A_1), \ldots, (f_k, Q_k, A_k) \}
\]
where $A_k$ is the final answer predicted by $P$, and $Q_i$ is a sub-query directed to the sub-task function $f_i \in F$. A high-level imperative controller manages the execution of $P$, passing inputs and outputs between the decomposer and sub-task handlers until the final output is obtained.

\begin{figure}[h]
  \centering
  \includegraphics[width=\linewidth]{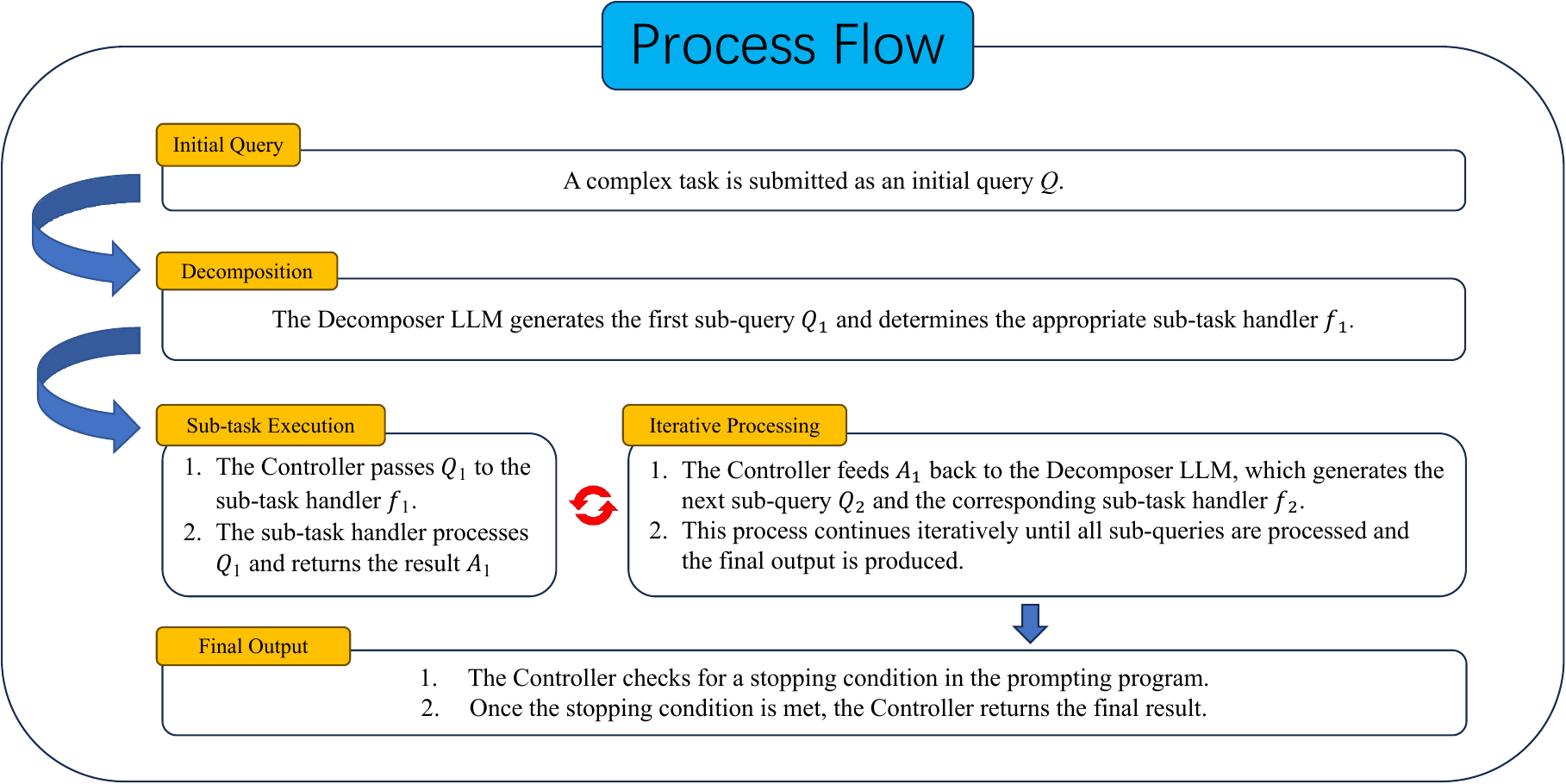}
  \caption{An example of the process flow of DECOMP.}
  \label{Process_Flow}
\end{figure}

To teach the decomposer LLM, in-context examples are used. These examples demonstrate the decomposition of complex queries into simpler sub-queries. Each example $E_j$ takes the form:
\[
E_j = \left( Q_j, \left\{ (f_{j,1}, Q_{j,1}, A_{j,1}), \ldots, (f_{j,k_j}, Q_{j,k_j}, A_{j,k_j}) \right\} \right)
\]
where $A_{j,k_j} = A_j$ is the final answer for $Q_j$, and $(Q_{j,1}, \ldots, Q_{j,k_j})$ represents the decomposition of $Q_j$. Each sub-task function $f$ is operationalized through sub-task handlers, which can be additional LLM prompts or symbolic or learned functions \cite{khot2023decomposedpromptingmodularapproach}. An illustration of the process flow is shown in Figure \ref{Process_Flow}.

The DECOMP approach has several advantages. First, its modularity allows each sub-task handler to be independently optimized, debugged, and upgraded, which facilitates systematic performance improvements and easier integration of new methods or models. Second, DECOMP can incorporate error-correcting sub-task handlers, improving the overall accuracy and reliability of the system. Third, the approach allows for diverse decomposition structures, including hierarchical and recursive decompositions, which are particularly useful for handling complex and large-scale problems. Finally, sub-task handlers can be shared across different tasks, enhancing the efficiency of the problem-solving process.

DECOMP and Least-to-Most Prompting \cite{zhou2023leasttomost} both decompose complex tasks to enhance large language models' reasoning abilities, but DECOMP distinguishes itself through its flexible, modular approach. Unlike Least-to-Most Prompting’s linear progression from easy to hard sub-questions, DECOMP allows for non-linear and recursive decomposition, with dedicated sub-task handlers that can be independently optimized and replaced. This modularity not only enhances flexibility and reusability across tasks but also introduces potential error-correcting mechanisms, making DECOMP more robust and adaptable to complex, multi-step reasoning tasks. While DECOMP has demonstrated superior performance in specific domains, such as symbolic reasoning and multi-step question answering, its advantages over Least-to-Most Prompting may vary depending on the nature of the task \cite{khot2023decomposedpromptingmodularapproach}.

In case studies, DECOMP demonstrated superior performance in various scenarios. For instance, in the k-th letter concatenation task, DECOMP outperformed CoT prompting by effectively teaching the sub-task of extracting the k-th letter through further decomposition. In list reversal, DECOMP showed better length generalization compared to CoT by recursively decomposing the task into reversing smaller sub-lists, achieving higher accuracy for longer input sequences. In long-context question answering (QA), DECOMP allowed for handling more examples than feasible with CoT prompting, leading to improved performance. In open-domain QA, incorporating symbolic retrieval APIs within the DECOMP framework enhanced performance on multi-hop QA datasets compared to CoT prompting. Additionally, in Math QA, DECOMP improved accuracy by post-processing CoT prompts to fix frequent formatting errors, resulting in significant performance gains \cite{khot2023decomposedpromptingmodularapproach}.

By leveraging the modular, flexible, and systematic approach of DECOMP, complex tasks can be effectively decomposed and solved, showcasing its superiority over traditional CoT prompting and other contemporary methods.

\subsection{Active prompt}
\begin{figure}[h]
  \centering
  \includegraphics[width=\linewidth]{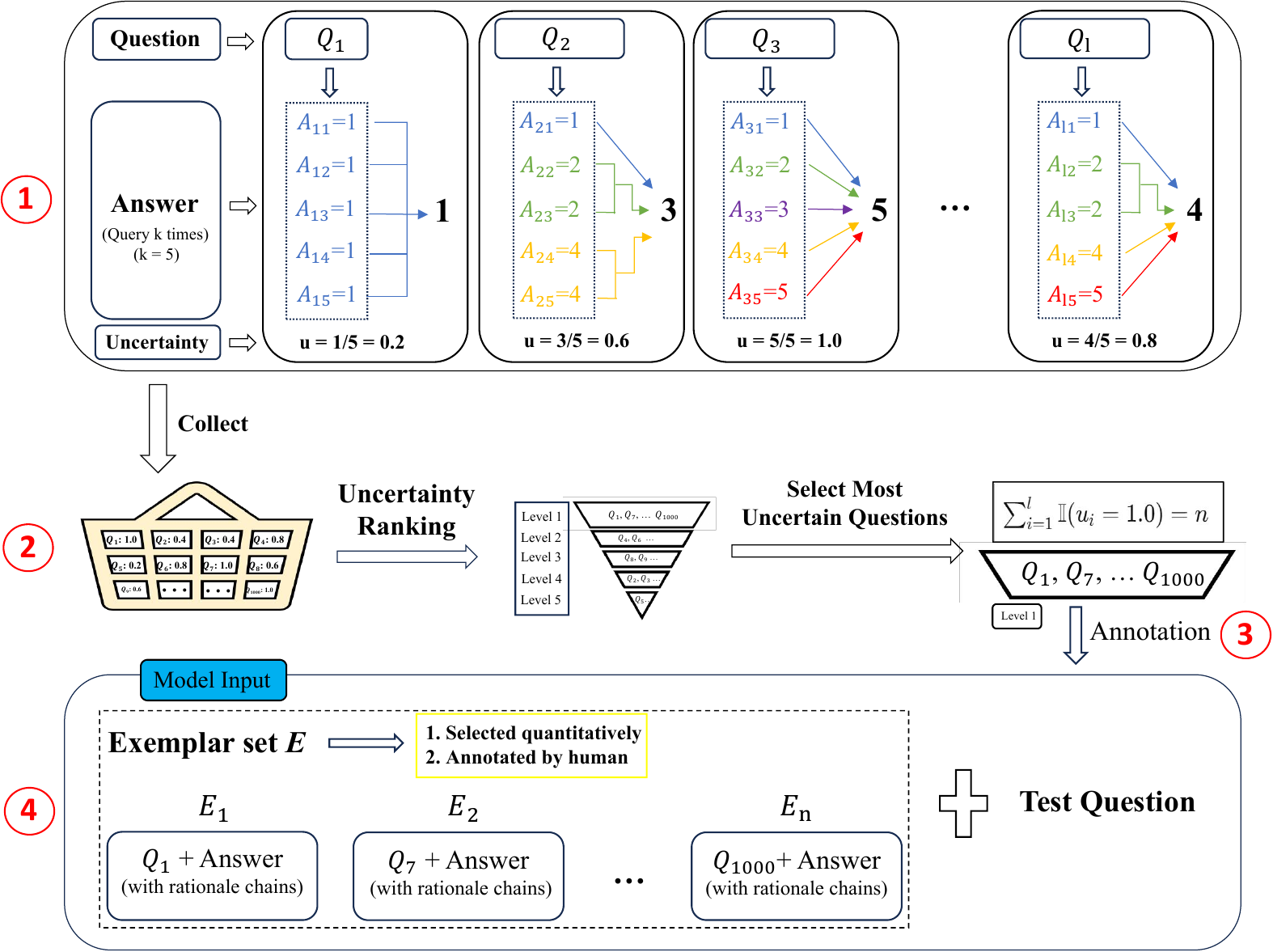}
  \caption{Illustration of the whole process. (1) Uncertainty Estimation. (2) Collection, Ranking and Selection. (3) Annotation (by human). (4) Inference.}
  \label{Active prompt}
\end{figure}

The adoption of the active prompt \cite{diao2024activepromptingchainofthoughtlarge} method marks a significant advancement in the utilization of LLMs for complex reasoning tasks. The active prompt method does not involve the traditional process of prefix-tuning \cite{li-liang-2021-prefix}. Instead, it focuses on improving the reasoning capabilities of LLMs through strategic selection and annotation of task-specific examples. By systematically selecting and annotating the most uncertain questions, this method not only refines the model's understanding but also leverages human expertise more effectively \cite{sahoo2024systematicsurveypromptengineering}. The process begins with the generation of multiple predictions for each question, followed by the calculation of uncertainty (uncertainty estimation) \cite{settles2009active, 10.5555/1619410.1619452} using various metrics such as disagreement, entropy, and variance. This strategic selection process ensures that the most informative questions are prioritized for annotation. The human annotation phase is crucial, as it involves providing detailed chain-of-thought reasoning and answers, which are then used to prompt the LLM during inference. This annotated data serves as exemplars, guiding the model through complex reasoning pathways and enhancing its predictive accuracy. The application of self-consistency \cite{wang2023selfconsistency} techniques further solidifies the model’s reliability by selecting the most consistent answers from multiple reasoning paths. The key innovation of this method is the thought of finding out the most efficient one-shot or few-shot \cite{logan-iv-etal-2022-cutting} examples, so it improves the inference ability of specific fields. A concrete process illustration is shown in Figure \ref{Active prompt}.

The active prompt method offers several key benefits, including efficient task adaptation and significant performance improvements across various reasoning domains. This approach aligns with the broader trend towards more interactive and adaptive AI systems, emphasizing the importance of responsive design in prompt engineering. Its ability to reduce human engineering efforts by focusing on the most uncertain and informative questions makes it an important tool for advancing LLM capabilities. This method not only enhances the quality of task-specific prompts but also maximizes the use of human expertise, paving the way for more sophisticated and accurate AI systems \cite{diao2024activepromptingchainofthoughtlarge}.

\subsection{Prompt pattern catalog}
A Prompt Pattern Catalog \cite{white2023prompt} is an organized collection of prompt templates and patterns designed to enhance the effectiveness of prompt engineering, particularly for LLMs such as ChatGPT. This methodology involves creating a standardized set of prompt patterns that can be applied across various tasks, ensuring consistency and optimizing the performance of models through systematic prompt design. By developing a catalog of prompt patterns, researchers and practitioners can ensure a consistent approach to prompt engineering, reducing variability and errors from ad hoc prompt creation \cite{white2023prompt, desmond2024exploring}. Predefined prompt patterns streamline the process of prompt engineering, saving time and resources by allowing practitioners to select and adapt patterns rather than crafting new prompts from scratch. A well-designed prompt pattern catalog includes patterns for various contexts and applications, enabling models to be quickly adapted to new tasks and domains by selecting the most appropriate patterns. Systematic use of optimized prompt patterns enhances model performance by providing more effective and contextually appropriate prompts, leading to better task-specific results \cite{mondal2024enhancing}.

The central methodology of this research involves the conceptualization and application of prompt patterns, which are reusable solutions to common problems encountered when interacting with LLMs. These prompt patterns are analogous to design patterns in software engineering, providing structured and documented approaches to enhance the output and interaction quality of LLMs. The framework for documenting these prompt patterns includes a detailed structure that ensures their adaptability across different domains \cite{white2023prompt}.

To systematically categorize these prompt patterns, the authors have divided them into five primary categories: Input Semantics, Output Customization, Error Identification, Prompt Improvement, and Interaction. This classification helps in organizing the patterns based on their functional roles and the specific problems they address. Within this framework, the research introduces a comprehensive catalog of 16 distinct prompt patterns. Each pattern is meticulously documented with the following components: name and classification, intent and context, motivation, structure and key ideas, example implementation, and practical consequences. The prompt patterns cover a wide range of functionalities. For instance, the Input Semantics category includes patterns such as Meta Language Creation, which helps in defining custom input languages for LLMs. The Output Customization category features patterns such as Output Automater and Visualization Generator, which tailor the generated outputs to specific formats or visualizations. Error Identification patterns such as Fact Check List ensure the accuracy of generated content by highlighting critical facts for verification. Prompt Improvement patterns, including Question Refinement and Alternative Approaches, enhance the quality of interactions by refining questions and suggesting multiple ways to achieve a goal. Lastly, Interaction patterns such as Flipped Interaction and Game Play facilitate dynamic and engaging user-LLM interactions \cite{white2023prompt}.

The methodology also emphasizes the combinatory use of these patterns to tackle more complex prompt engineering tasks. By providing detailed examples and practical implementations, the research demonstrates how multiple prompt patterns can be integrated to create sophisticated and efficient prompting strategies. This structured approach not only improves the effectiveness of LLMs in various applications but also contributes to the broader understanding and advancement of prompt engineering as a field \cite{white2023prompt}.

Research supports the effectiveness of prompt pattern catalogs. \cite{white2023prompt} outlines that the development and use of a prompt pattern catalog can improve the effectiveness and efficiency of prompt engineering with LLMs. \cite{mondal2024enhancing} explores how predefined structured prompt patterns can enhance user interaction and improve model outputs in conversational AI. \cite{desmond2024exploring} investigates the application of prompt engineering patterns in enterprise settings, demonstrating their utility in optimizing model performance across various tasks. Additionally, \cite{white2023chatgpt} highlights the benefits of using predefined structured prompt patterns in software development, demonstrating significant improvements in code quality, requirements elicitation, and refactoring efficiency.

\subsection{Prompt optimization}
In the domain of prompt engineering for LLMs, the challenge of crafting effective prompts remains a significant barrier due to the extensive manual effort and expertise required. Prompt optimization is a critical technique for improving the performance of LLMs by refining the input prompts that guide their responses. The process of prompt optimization systematically adjusts these prompts to enhance accuracy and relevance, reducing the need for manual trial and error.

Several methods have been developed to automate prompt optimization, including gradient-based approaches such as Prompt Optimization with Textual Gradients (ProTeGi) \cite{pryzant-etal-2023-automatic}, which uses text-based gradients to iteratively refine prompts, and black-box methods that optimize prompts based solely on output performance without requiring model internals. Additionally, model-adaptive techniques, such as Model-Adaptive Prompt Optimization (MAPO) \cite{chen-etal-2023-mapo}, tailor the optimization to the specific characteristics of the LLM, potentially offering superior results. Each method has its advantages: gradient-based techniques are efficient and directed, black-box approaches are broadly applicable and easy to implement, and model-adaptive methods provide customized optimization for specific models. The choice of method depends on task requirements, model complexity, and available resources. 

\subsubsection{Prompt optimization with textual gradients}
Prompt Optimization with Textual Gradients (ProTeGi) \cite{pryzant-etal-2023-automatic} is inspired by gradient descent, a fundamental technique in optimization, but adapts this concept to the discrete and non-parametric nature of natural language processing. Instead of relying on numerical gradients, ProTeGi generates ``textual gradients'', which are natural language descriptions of the flaws in a given prompt based on its performance on a small batch of data. These gradients indicate the semantic direction in which the prompt needs to be improved.

ProTeGi further enhances this optimization process by applying these textual gradients to modify the prompt in the opposite semantic direction, akin to a reverse gradient descent in the language space. This iterative process is guided by a beam search algorithm combined with a bandit selection strategy, which efficiently explores the space of possible prompts and selects the most promising candidates for further refinement \cite{pryzant-etal-2023-automatic}.

The effectiveness of ProTeGi has been demonstrated across multiple NLP tasks, including sentiment analysis, fake news detection, and the novel problem of LLM jailbreak detection. Experimental results indicate that ProTeGi can significantly improve prompt performance, with reported gains of up to 31\% over initial prompts, while also surpassing existing prompt optimization methods in efficiency and accuracy. This method provides a robust, data-driven approach to prompt engineering, offering a scalable solution that can adapt to various tasks without requiring access to the internal states of LLMs \cite{pryzant-etal-2023-automatic}.

\subsubsection{Black-box prompt optimization}
In recent prompt engineering research, the challenge of aligning LLMs with human intent without model retraining has garnered significant attention. Traditional alignment techniques, such as Reinforcement Learning from Human Feedback (RLHF) and Direct Preference Optimization (DPO), typically require substantial computational resources and direct access to model parameters, which are not always feasible or efficient, particularly with closed-source models such as GPT-4 or Claude-2. In response to these limitations, a novel method called Black-box Prompt Optimization (BPO) \cite{cheng-etal-2024-black} has been introduced, providing a promising alternative for enhancing model alignment through prompt optimization alone.

BPO shifts the focus from model-centric to input-centric optimization, where the key idea is to refine the user's prompts rather than altering the model's internal parameters. This approach leverages feedback from pre-existing datasets that contain human preferences, creating pairs of original and optimized prompts. These pairs are then used to train a sequence-to-sequence model designed to rewrite prompts in a way that improves the alignment of LLM outputs with human expectations \cite{cheng-etal-2024-black}.

The BPO method offers several advantages. First, it is model-agnostic, allowing it to be applied across various LLMs, whether open-source or API-based, without requiring access to the model’s internals. Second, it enhances interpretability, as the changes made to prompts are transparent and directly observable, providing clear insights into how and why a particular prompt leads to better alignment. Third, empirical results demonstrate that BPO not only improves the alignment of models such as GPT-3.5 and LLaMA-2 but also outperforms RLHF and DPO when used independently or in conjunction with these methods \cite{cheng-etal-2024-black}.


\subsubsection{Model-adaptive prompt optimization}
Traditionally, prompt optimization has focused on tailoring prompts to specific tasks to enhance model performance. However, \cite{chen-etal-2023-mapo} highlighted the necessity of adapting prompts not just to tasks but also to the specific characteristics of different LLMs. This shift in perspective has led to the development of Model-Adaptive Prompt Optimization (MAPO), a novel approach designed to fine-tune prompts for individual LLMs, thereby maximizing their effectiveness across various downstream tasks. MAPO addresses the inherent variability in how different LLMs respond to the same prompt by introducing a two-phase optimization process. The first phase involves establishing a warm-up dataset, where candidate prompts are generated and evaluated for their suitability to each LLMs. This is followed by a combination of Supervised Fine-Tuning (SFT) and Reinforcement Learning (RL), particularly employing techniques such as Proximal Policy Optimization (PPO) and Ranking Responses from Model Feedback (RRMF). This joint learning approach refines the prompts, ensuring they align with the specific preferences of each LLMs.

Empirical studies demonstrate that MAPO significantly improves performance in tasks such as question-answering, classification, and text generation when compared to conventional task-specific prompt optimization methods. By shifting the focus from a one-size-fits-all approach to a more nuanced, model-specific strategy, MAPO enhances the robustness and generalization of LLMs, making it a powerful tool in the prompt engineering toolkit \cite{chen-etal-2023-mapo}.

\subsubsection{PromptAgent}
The PromptAgent method suggests framing prompt optimization as a strategic planning problem. A kind core of this method is the use of Monte Carlo Tree Search (MCTS), a principled planning algorithm that strategically navigates the vast space of expert-level prompts. Unlike conventional methods that generate prompts through local variations, PromptAgent employs a trial-and-error mechanism, inspired by human problem-solving strategies. This approach allows the model to iteratively refine prompts based on error feedback, simulating future rewards and prioritizing high-reward paths \cite{wang2023promptagentstrategicplanninglanguage}.
Another core of this method, for instance, PromptSource \cite{bach2022promptsource}, collects over 2,000 open-source prompts for roughly 170 datasets, by dataset exploration, prompt writing and documentation to provide an enhanced prompt. 

PromptAgent's effectiveness has been demonstrated across a diverse set of tasks, spanning general NLP challenges and domain-specific applications such as biomedical text processing. By autonomously generating prompts that incorporate domain-specific knowledge and detailed task instructions, PromptAgent consistently outperforms both human-designed prompts and other automated optimization methods \cite{wang2023promptagentstrategicplanninglanguage}, highlighting the importance of integrating strategic planning and self-reflection capabilities into prompt optimization frameworks. 

\subsubsection{Reinforcement learning}
Reinforcement Learning (RL) for prompt optimization is an advanced technique designed to enhance the performance of LLMs by iteratively refining the prompts used during training and inference. This method utilizes the principles of reinforcement learning to navigate the complex parameter space of large models, optimizing the prompts for improved task-specific performance. In RL for prompt optimization, a reward function is defined to evaluate the effectiveness of different prompts based on the model's output. The model then uses this feedback to adjust and optimize the prompts through a series of iterations, ensuring that the prompts evolve to maximize performance on the target task by leveraging the model's ability to learn from its interactions with the environment \cite{deng-etal-2022-rlprompt}.

Consider the task of VQA, where the goal is to generate accurate answers to questions based on visual input. Using RL for prompt optimization, the model can start with a set of initial prompts and iteratively refine them based on the accuracy of the generated answers. For instance, if the model is asked, ``What is the color of the car in the image?" the initial prompts might produce varied responses. The reward function will assess these responses, favoring prompts that lead to correct answers. Over multiple iterations, the model learns to generate more precise prompts, improving its ability to accurately answer similar questions in the future \cite{awal2024investigating}.

\subsubsection{GPTs (plugins)}
\begin{figure}[h]
  \centering
  \includegraphics[width=\linewidth]{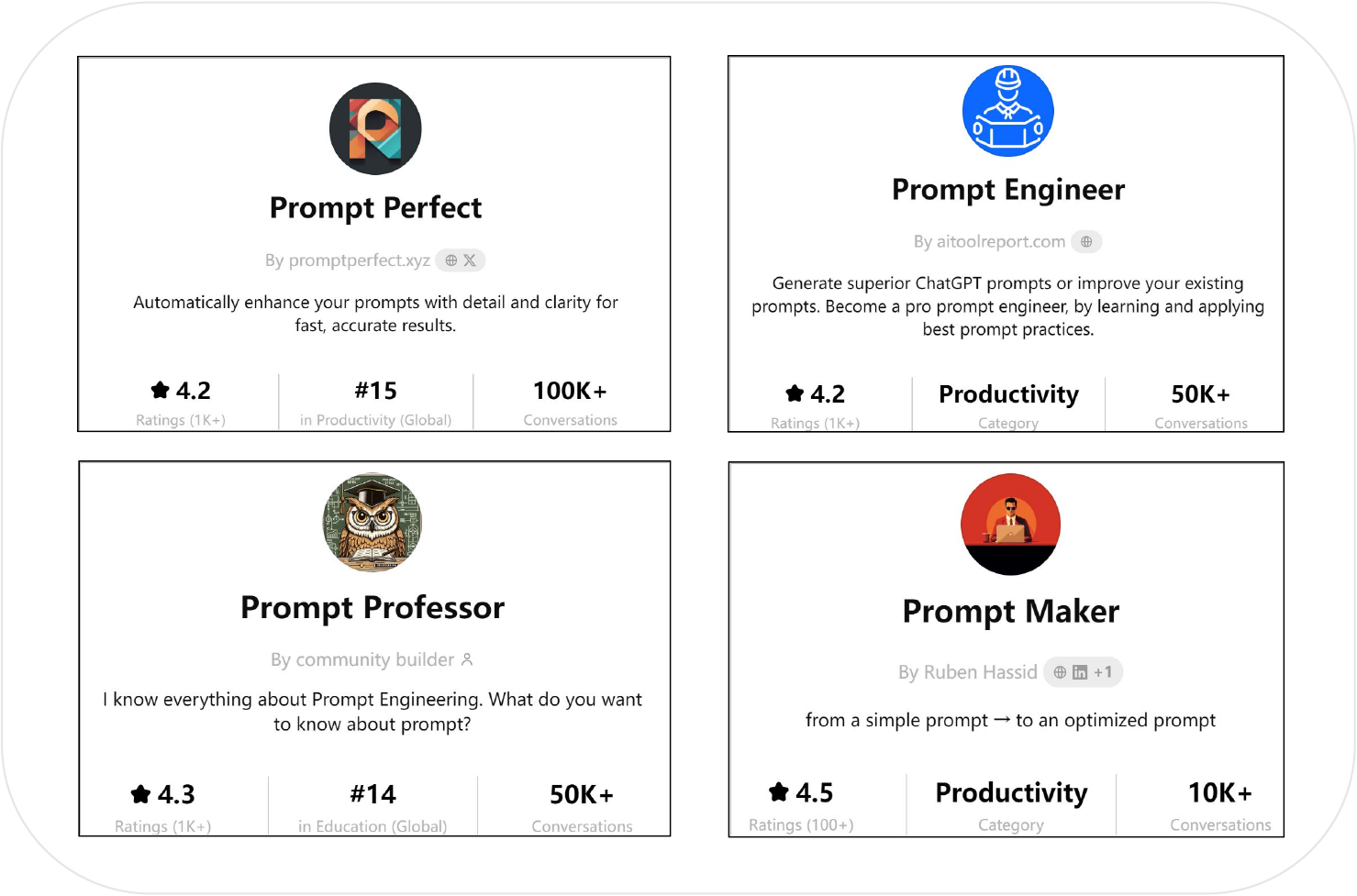}
  \caption{Examples of GPT apps that specialize in generating or optimizing prompts \cite{ChatGPT2024}.}
  \label{GPTs(plugins)}
\end{figure}

Before ending this discussion on prompt optimization techniques, we need to mention the use of external prompt engineering assistants that have been developed recently and exhibit promising potential. Unlike the methods introduced previously, these instruments can help us to polish the prompt directly. They are adept at analyzing user inputs and subsequently producing pertinent outputs within a context that is defined by itself, thereby amplifying the efficacy of prompts. Some of the plugins provided by the OpenAI GPT store are good examples of such tools \cite{openai2023chatgptplugins}. Some popular GPT store apps that specialize in generating or optimizing prompts are shown in Figure \ref{GPTs(plugins)}.

In certain implementations, the definition of a plugin is incorporated into the prompt, altering the output \cite{bisson2023microsoft}. Such integration may impact the manner in which LLMs interpret and react to the prompts, illustrating a connection between prompt engineering and plugins. Plugins mitigate the laborious nature of intricate prompt engineering,  enabling the model to more proficiently comprehend or address user inquiries without necessitating excessively detailed prompts. Consequently, plugins can bolster the efficacy of prompt engineering while promoting enhanced user-centric efficiency. These tools, akin to packages, can be seamlessly integrated into Python and invoked directly \cite{ng2023prompt,roller2021recipes}. 
For instance, the ``Prompt Enhancer" pluging \cite{whatplugin2023promptenhancer}, developed by AISEO \cite{aiseo2023}, can be invoked by starting the prompt with the word ``AISEO" to let the AISEO prompt generator automatically enhance the LLM prompt provided. Similarly, another plugin called ``Prompt Perfect", can be used by starting the prompt with `perfect' to automatically enhance the prompt, aiming for the ``perfect" prompt for the task at hand \cite{chatonai2023promptperfect, promptperfect2023terms}. Nevertheless, while the use of plugins to enhance prompts is simple and handy, it is not always clear which prompt engineering technique, or combination of techniques, is implemented by a given plugin, given the closed-source nature of most plugins.

\subsection{Retrieval augmentation}
Another direction of prompt engineering research is to aim to reduce hallucinations. When using AIGC tools such as GPT-4, it is common to face a problem called ``hallucinations", which refer to the presence of unreal or inaccurate information in the model's generated output \cite{lee2018hallucinations,maynez2020faithfulness}. While these outputs may be grammatically correct, they can be inconsistent with facts or lack real-world data support. Hallucinations arise because the model may not have found sufficient evidence in its training data to support its responses, or it may overly generalize certain patterns when attempting to generate fluent and coherent output \cite{ji2023hallu}. 

An approach to reduce hallucinations and enhance the effectiveness of prompts is the so-called retrieval augmentation technique, which aims at incorporating \mbox{up-to-date} external knowledge into the model's input \cite{lazaridou2022internet,jiang2023active}. It is emerging as an AI framework for retrieving facts from external sources. \cite{ram2023incontext} examines the augmentation of context retrieval through the incorporation of external information. It proposes a sophisticated operation: the direct concatenation of pertinent information obtained from an external source to the prompt, which is subsequently treated as foundational knowledge for input into the expansive language model. Additionally, the paper introduces auto-regressive techniques for both retrieval and decoding, facilitating a more nuanced approach to information retrieval and fusion. \cite{ram2023incontext} demonstrates that in-context retrieval-augmented language models, when constructed upon readily available general-purpose retrievers, yield significant LLM enhancements across a variety of model dimensions and diverse corpora. In another research, \cite{shuster2021retrieval} showed that GPT-3 can reduce hallucinations by studying various implementations of the retrieval augmentation concept, such as Retrieval Augmented Generation (RAG) \cite{lewis2020rag}, Fusion-in-Decoder (FiD) \cite{izacard2020fid}, Seq2seq \cite{lewis2020bart,raffel2020t5,roller2021blender} and others. \cite{dhuliawala2023chain} developed the Chain-of-Verification (CoVe) approach to reduce hallucinations, based on letting the LLM deliberate on its own responses before self-correcting them. They suspect that extending this approach with retrieval augmentation would likely bring further gains. UNIWEB \cite{2305109982023web} converting knowledge-intensive tasks into a unified text-to-text framework and treating the web as a general source of knowledge.

\subsection{Reasoning and active interaction}
This subsection explores two advanced techniques that enhance the capabilities of LLMs by integrating reasoning with interaction through external tools or other action abilities. Automatic Reasoning and Tool Usage (ART) combines CoT prompting with the use of specialized tools. By guiding LLMs through multi-step reasoning and incorporating resources such as calculators and databases, ART improves the logical coherence and accuracy of model outputs. The ReAct Framework (Reasoning and Acting) synergizes reasoning with actionable steps. It prompts LLMs to devise logical sequences and interact dynamically with external tools, enabling them to handle complex, multi-step tasks efficiently. Both ART and ReAct represent significant advancements in prompt engineering, enhancing the range and reliability of tasks that LLMs can perform through the integration of reasoning and interaction.

\subsubsection{Automatic reasoning and tool usage}
ART is an advanced prompting technique that combines the principles of automatic  CoT prompting with the strategic utilization of external tools. This method aims to enhance the reasoning capabilities of LLMs by guiding them through multi-step reasoning processes and leveraging specialized tools to achieve more accurate and relevant outputs \cite{paranjape2023art}.

ART builds on the CoT prompting technique, which encourages models to generate intermediate reasoning steps before arriving at a final answer. In ART, these reasoning steps are augmented by incorporating external tools such as calculators, databases, or other software applications. The integration of tools helps LLMs to perform tasks that require precise calculations, access to updated information, or specialized data processing that the model alone may not handle effectively.

For example, a prompt designed using ART might guide an LLM to first outline the steps required to solve a complex mathematical problem and then use a calculator tool to perform the necessary calculations. This combination of reasoning and tool usage ensures that the model's outputs are both logically coherent and computationally accurate.

\cite{Greyling2023} have demonstrated that ART can help models navigate complex problem spaces more effectively by breaking down tasks into manageable steps and utilizing appropriate tools at each stage. For instance, the integration of ART in natural language processing tasks has shown promising results in areas such as automated customer service, where models need to access and process information dynamically \cite{Vinija2023}.

Moreover, ART's approach aligns with ongoing efforts to develop more robust and versatile AI systems capable of handling real-world tasks that demand a combination of cognitive and computational skills. \cite{Badhan2023} explores advanced ART techniques to achieve better accuracy and reliability in AI applications. These findings underscore the importance of ART in enhancing the functionality and performance of LLMs, making them more adept at handling a broader range of tasks, in particular technical problem-solving tasks that require specific and precise outputs such as financial calculations or data analysis.

\subsubsection{ReAct framework}

The ReAct Framework, which stands for Reasoning and Acting, synergizes the processes of reasoning and action to enable LLMs to not only think through problems but also interact with external tools and environments to achieve more accurate and contextually appropriate outcomes.

The ReAct Framework operates by prompting LLMs to generate both reasoning traces and task-specific actions. This dual approach ensures that the model first contemplates the problem, devises a logical sequence of thoughts, and then executes actions that may involve querying external databases, using calculators, or interacting with other software tools. This method is particularly effective in scenarios requiring detailed reasoning followed by specific actions, thus ensuring the LLM can handle complex, multi-step tasks efficiently \cite{yao2023react}.

\begin{figure}[h]
  \centering
  \includegraphics[width=\linewidth]{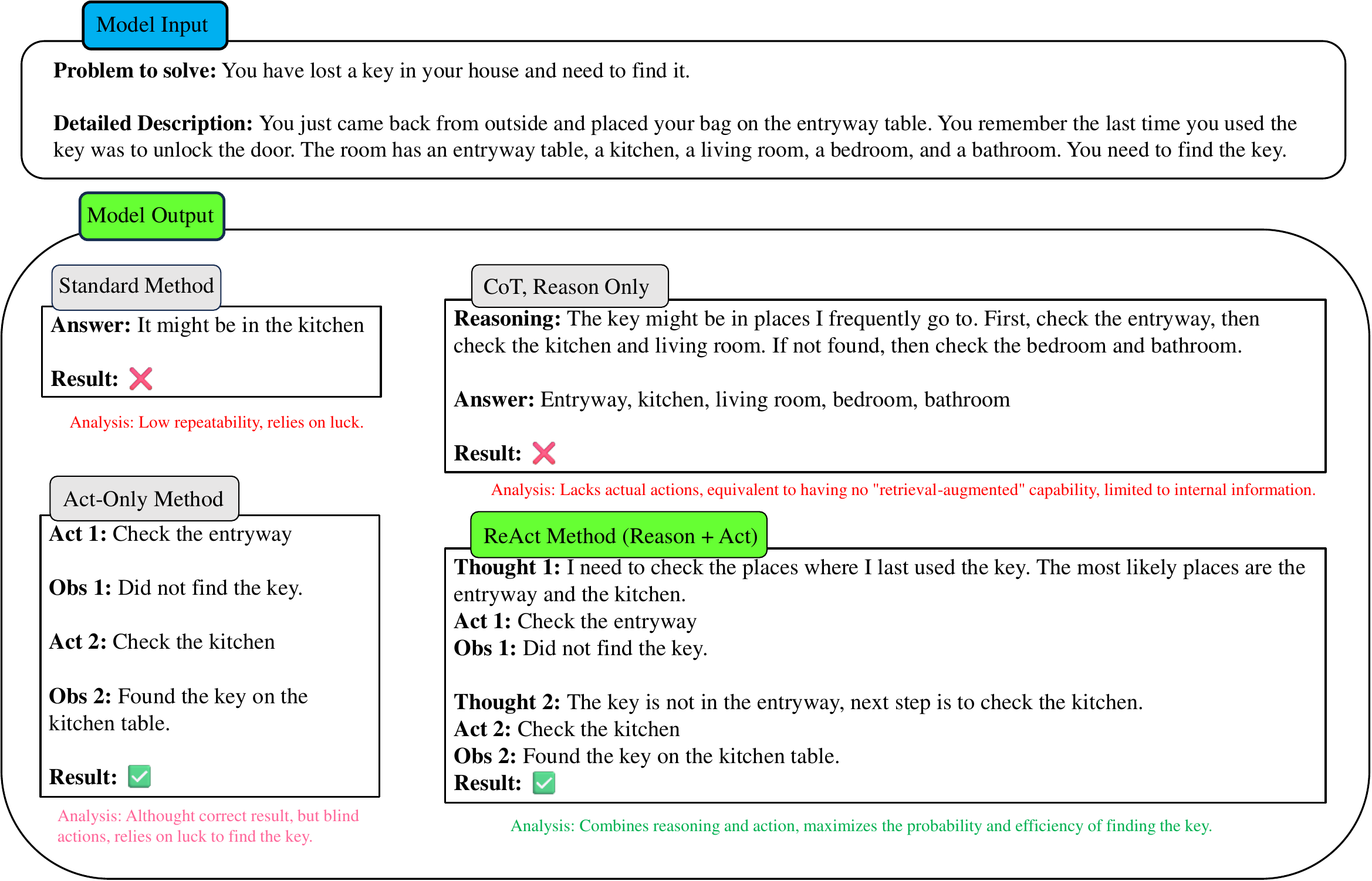}
  \caption{An example of ReAct method.}
  \label{Reason_and_Act}
\end{figure}

For example, in a task involving financial analysis, the ReAct framework would first prompt the LLM to outline the necessary steps to evaluate a portfolio. Subsequently, the model could use financial analysis tools to gather current market data and perform calculations, integrating these results into the final analysis. This combination of reasoning and action leads to more robust and reliable outcomes compared to using static prompts alone. Another concrete example is shown in Figure \ref{Reason_and_Act}.

By integrating reasoning and action, ReAct enables LLMs to make more informed and accurate decisions, in particular in fields such as finance, healthcare, and legal analysis, where decisions must be based on comprehensive data and logical reasoning \cite{Li2023}. Additionally, the framework's ability to interact with external tools ensures that the information used in decision-making is up-to-date and relevant, reducing the likelihood of errors due to outdated or incomplete data \cite{Roberts2023}. 

Implementing the ReAct framework is not a trivial task, as it involves developing prompts that guide LLMs through both thought processes and actions. This requires a detailed understanding of the task at hand and the tools available, ensuring that the model can seamlessly transition from reasoning to action. 

\section{Methodologies for multimodal large models}
\label{Sec4}
In recent years, vision-language models (VLMs) have made significant advancements in multimodal learning by combining visual and linguistic information. These models have demonstrated strong capabilities in tasks such as image description and visual question answering (VQA) \cite{Antol_2015_ICCV, WU201721, 8046084, KAFLE20173}. While this review primarily focuses on the potential of prompt engineering in LLMs, it is also pertinent to briefly introduce the importance of VLMs and their applications in multimodal tasks to provide a more comprehensive perspective.

VLMs are based on the transformer architecture, and are trained on extensive datasets to learn complex semantic relationships. However, unlike early unimodal models, VLMs process both textual and visual information, enabling them to establish connections between image understanding and text generation. As can be expected, this multimodal integration makes VLMs particularly effective at handling complex tasks that involve both images and text.

To seamlessly integrate and interpret these diverse data types, VLMs require sophisticated prompt designs that ensure contextual coherence and accuracy \cite{yin2024survey, 10386743}. Challenges such as data alignment, modality integration, and context preservation are addressed through advanced techniques such as Context Optimization (CoOp, subsection \ref{subsec:context_optimization}) and Multimodal Prompt Learning (MaPLe, subsection \ref{subsec:multimodal_prompt_learning}). These advanced prompt engineering techniques enhance the ability of VLMs to generate nuanced and contextually rich outputs, thereby facilitating their effective utilization in various applications and enabling them to tackle more complex tasks \cite{yin2024survey}.

\subsection{Zero-shot and few-shot prompting}
Zero-shot and few-shot prompting, which have already been discussed in subsection \ref{subsec:one-shot-few-shot} in the context of LLMs, are also pivotal techniques in the realm of VLMs, enabling these models to handle tasks with minimal or no task-specific training data. Zero-shot prompting allows models to perform tasks without any specific examples provided during training, relying entirely on their pre-trained knowledge to generalize across new tasks and domains. For example, a model such as CLIP can be prompted with a textual description to classify images into categories it has never explicitly been trained on \cite{brownGPT3}. On the other hand, few-shot prompting involves providing the model with a small number of examples during inference, significantly enhancing the model’s ability to generalize with limited data~\cite{Radford2021LearningTV}.

In relation to these methods, \cite{awal2024investigating} systematically explored a range of prompting techniques for zero-shot and few-shot visual question answering (VQA) in vision-language models (VLMs), highlighting the impact of question templates, the integration of image captions, and the application of chain-of-thought reasoning on model performance. \cite{Radford2021LearningTV} showed the application of these techniques in CLIP, highlighting the model’s ability to generalize across different domains. Additionally, \cite{zhang2022tipadapter} presented a method for adapting CLIP to few-shot classification tasks without additional training, emphasizing practical benefits in real-world applications.

\subsection{Continuous prompt vectors}
Advancements in prompt engineering have enabled more effective adaptation of pre-trained VLMs to a wide range of downstream tasks. A promising approach in this domain is the use of continuous prompt vectors to fine-tune models such as CLIP for complex video understanding tasks. Unlike traditional handcrafted prompts, which require expert knowledge and manual effort, continuous prompt vectors \cite{10.1007/978-3-031-19833-5_7} are learned during the training process, allowing for more flexible and efficient model adaptation. This method involves appending or prepending sequences of random vectors to the input text, which the model then interprets as part of its textual input. These vectors are optimized to effectively bridge the gap between the static image-based pre-training objectives and the dynamic requirements of video tasks, such as action recognition, action localization, and text-video retrieval. Additionally, lightweight temporal modeling using Transformers is applied to capture the temporal dependencies inherent in video data.

The efficiency of this approach lies in its minimal computational requirements; only a few parameters are trained, while the core model remains frozen. Despite this, the method has demonstrated competitive performance across various benchmarks, highlighting its potential in extending the capabilities of VLMs to handle resource-intensive video tasks with greater flexibility and accuracy. This continuous prompt-based adaptation represents a significant step forward in the evolution of prompt engineering, offering a scalable and effective solution for leveraging pre-trained models in more complex and diverse applications \cite{10.1007/978-3-031-19833-5_7}.

\subsection{Context optimization\label{subsec:context_optimization}}
Context Optimization (CoOp) \cite{Zhou_2022} is an innovative prompt learning approach specifically designed for VLMs. CoOp focuses on enhancing the adaptability and performance of these models by optimizing context-specific prompts. This methodology involves the creation of learnable context vectors that are embedded within the model's architecture, enabling it to dynamically adjust to different downstream tasks.

CoOp leverages the dual-stream architecture of VLMs, such as CLIP \cite{Radford2021LearningTV} and ALIGN \cite{pmlr-v139-jia21b}, by performing context optimization on top of these pre-trained models. CoOp introduces learnable context vectors that are fine-tuned to minimize classification loss, thus avoiding extensive manual prompt engineering. By utilizing learnable context vectors, CoOp fine-tunes the prompts to align with the specific characteristics of the complex input data. This results in improved performance and better generalization across various scenarios \cite{ma2023understanding}. This method is particularly valuable in applications such as image recognition and VQA, where the context can vary significantly \cite{agnolucci2023eco}.

To illustrate the practical application of CoOp, consider a VQA task \cite{Antol_2015_ICCV, WU201721, 8046084, KAFLE20173}. In a VQA scenario, the model is presented with an image and a corresponding question, and it must generate an accurate answer based on the visual and textual information. By leveraging CoOp, the model utilizes learnable context vectors to optimize the prompts specific to the context of the input image and question. This process enhances the model's ability to interpret the visual elements and comprehend the textual query, leading to more precise and contextually relevant answers. For instance, if the model is shown an image of a beach scene with the question ``What activity are the people engaged in?", CoOp would utilize learnable context vectors to optimize the textual prompts. These context vectors help the text encoder generate features that focus on relevant aspects of the image, such as identifying people, recognizing activities, and understanding the overall context of the scene. By aligning these optimized text features with the image features extracted by the image encoder, CoOp enables the model to generate a precise and contextually relevant answer, such as ``The people are playing volleyball on the beach."

Regarding CoOp's effectiveness, \cite{Zhou_2022} showed that models using CoOp significantly outperform traditional models in tasks such as image recognition and VQA. Additionally, \cite{agnolucci2023eco} highlighted the benefits of ensembling context optimization, which further enhances the model's performance by combining multiple context vectors. This approach has been shown to improve the robustness and generalization of VLMs in real-world applications \cite{chowdhury-etal-2023-apollo}.

\subsection{Conditional prompt learning}
Conditional Context Optimization (CoCoOp) \cite{zhou2022conditional} is a methodology that dynamically tailors prompts based on specific conditions or contexts. Specifically, CoCoOp employs a lightweight neural network to generate input-conditional prompt vectors for each image, ensuring that the pre-trained model parameters remain unchanged. By leveraging contextual information, CoCoOp can provide more precise and relevant guidance to the model, which is particularly useful in complex, multimodal scenarios where the interplay between different types of data must be carefully managed.

One significant advantage of CoCoOp is its ability to adapt to new and unseen data without the need for fine-tuning the pre-trained model, thanks to the context-specific prompts generated by the lightweight neural network. In other words, a VLM enhanced with conditional prompts can more accurately interpret and respond to images and questions it has not encountered during training \cite{zhou2022conditional,ma2024understanding}. This capability is critical for applications such as image captioning, VQA, and scene understanding, where the context can vary widely.

Consider an image captioning task where the goal is to generate descriptive captions for images. Using CoCoOp, the model enhances its performance with dynamically generated prompts tailored for different types of scenes. Specifically, CoCoOp extends the CoOp method by training a lightweight neural network to generate input-conditional tokens for each image. As detailed in \cite{zhou2022conditional}, this allows the model to adapt to various contexts without extensive retraining, resulting in more accurate and contextually relevant captions. For example, a prompt for an outdoor scene might include contextual cues related to nature, weather, and activities, while a prompt for an indoor scene might focus on objects, people, and interactions. For an image of a bustling market, the conditional prompt could include cues such as ``Identify the types of products being sold'' or ``Describe the interactions between vendors and customers''. This enables the model to produce a caption such as ``Vendors selling fresh fruits and vegetables in a crowded market, with customers browsing and purchasing items'' \cite{zhou2022conditional}. 

This dynamic adaptation improves caption accuracy and enhances the model's ability to generalize to novel scenes, addressing the limitations of static prompt methods such as CoOp. Besides image captioning, the improved generalization capabilities of this technique make the model more robust in tasks such as VQA, image classification, and other real-world applications \cite{khattak2023selfregulating}.

\subsection{Multimodal prompt learning\label{subsec:multimodal_prompt_learning}}

The core idea of Multimodal Prompt Learning (MaPLe) is to introduce and optimize prompts for both the vision and language components simultaneously. By embedding prompts at various stages within the transformer architecture, MaPLe ensures that the model can adaptively learn contextual information pertinent to the specific task at hand \cite{khattak2023maple}. This hierarchical approach allows the model to progressively refine its understanding and integration of multimodal inputs, leading to improved performance across a range of applications.

\begin{figure}[h]
  \centering
  \includegraphics[width=\linewidth]{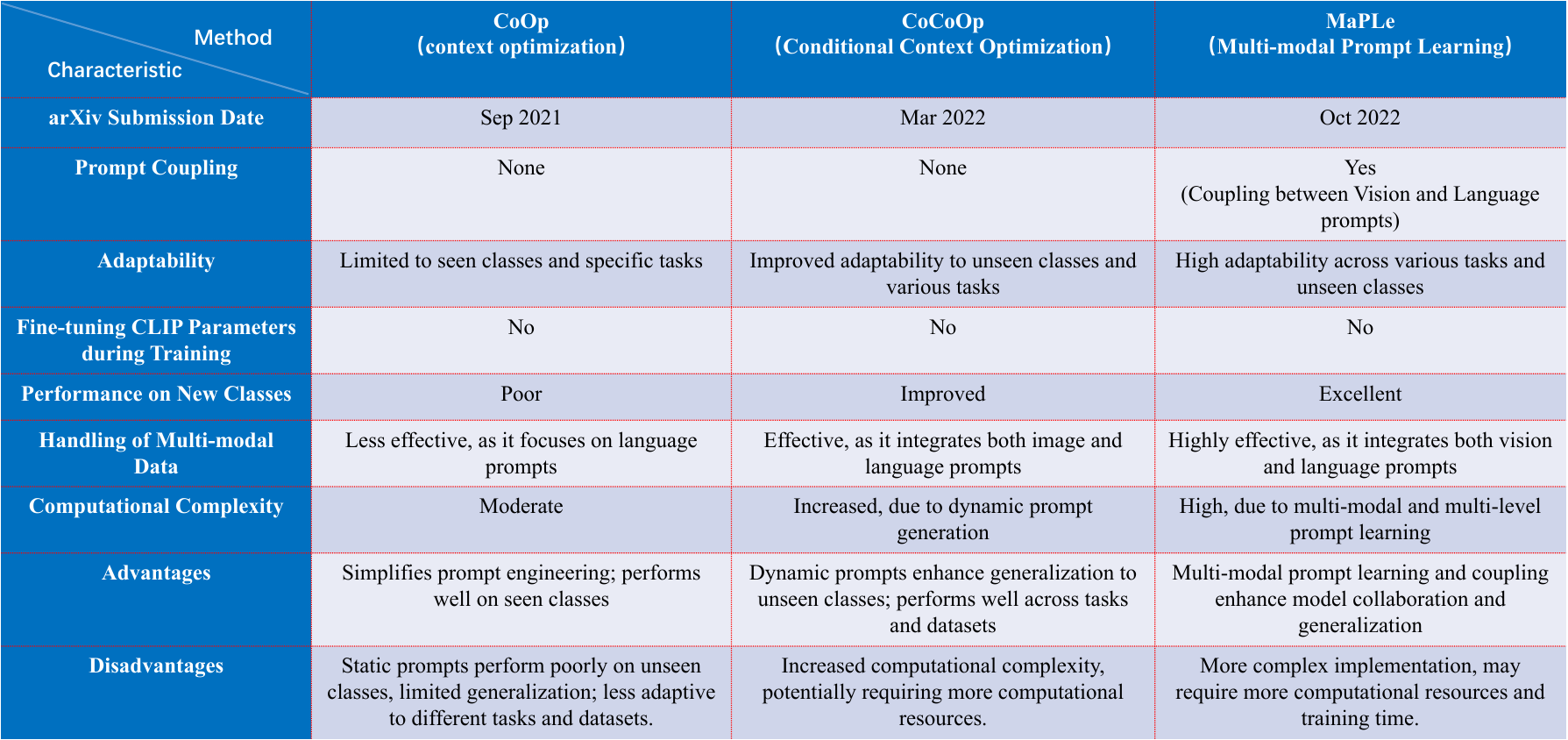}
  \caption{Comparison between the MaPLe and traditional method (CoOp and CoCoOp).}
  \label{MaPLe VS CoOp&CoCoOp}
\end{figure}

One of the critical innovations of MaPLe is its ability to enhance task relevance. Traditional prompt engineering often focuses on either vision or language prompts in isolation, which can limit the model's ability to fully leverage the complementary information available in multimodal data. MaPLe overcomes this limitation by jointly optimizing prompts for both modalities, thereby facilitating a more integrated and coherent representation of the input data \cite{khattak2023maple,ma2024understanding}. A detailed comparison between the MaPLe and traditional method (CoOp and CoCoOp) is shown in Figure \ref{MaPLe VS CoOp&CoCoOp}.

Another important mechanism of MaPLe, the hierarchical learning mechanism, allows the model to process and integrate information at multiple levels of abstraction. This is particularly beneficial for complex tasks that require a deep understanding of both visual and textual elements. By optimizing prompts at different layers within the transformer, MaPLe can better capture the intricate dependencies between vision and language inputs \cite{khattak2023maple,gu2023systematic}.

\cite{khattak2023maple} showed that MaPLe significantly outperforms baseline models in tasks such as image recognition and VQA. Similarly, \cite{ma2024understanding} highlighted the importance of Multimodal prompt learning in enhancing the adaptability and generalization of VLMs.

To illustrate the practical application of MaPLe, consider the task of VQA \cite{Antol_2015_ICCV, WU201721, 8046084, KAFLE20173}. In a typical VQA scenario, a model is provided with an image and a related question, and it must generate a correct and contextually relevant answer. Using MaPLe, the model can be fine-tuned with multimodal prompts that simultaneously address both the visual content and the textual question. For instance, given an image of a bustling market and the question ``What fruit is the vendor selling?", MaPLe would embed prompts at various levels of the transformer's vision and language branches. These prompts might include visual prompts that focus on identifying objects and text prompts that guide the model to look for specific answer-relevant details. By processing these prompts hierarchically, the model can effectively integrate visual cues (like recognizing apples and oranges in the image) with the textual context (understanding the question) to generate an accurate answer (e.g., ``The vendor is selling apples and oranges"). This multimodal approach ensures that the model leverages both the visual and textual information in a coherent and integrated manner, resulting in improved performance on VQA tasks compared to models that do not utilize such comprehensive prompt learning strategies.

\section{Assessing the efficacy of prompt methods}
\label{Sec5}

There exist several ways to evaluate the quality of the output of an LLM. Evaluation methods can generally be divided into subjective and objective categories to assess the efficacy of current prompt methods in AIGC tools

\subsection{Subjective and objective evaluations}
The task of prompt engineering can be challenging because it is difficult to determine how a prompt is more effective solely based on its raw text form \cite{2305107132023flatness}. Therefore, evaluating prompts requires a combination of subjective and objective methods. Subjective evaluations primarily rely on human evaluators to assess the quality of the generated content. Objective evaluations, also known as automatic evaluation methods, use algorithms to score the quality of text generated by LLMs or test on various benchmarks to quantitatively measure the efficacy of prompt methods. 

Subjective evaluation and objective evaluation methods each have their advantages and disadvantages. Subjective evaluation is more in line with human intuition, but it is also more expensive and time-consuming \cite{2304019042023refiner}. Objective evaluation is less expensive and quicker than subjective evaluation. For instance, despite numerous pieces of research highlighting the limited correlation between BLEU and alternative metrics based on human assessments, their popularity has remained unaltered \cite{ananthakrishnan2007some,callison2006re}.
The best way to evaluate the quality of LLM output depends on the specific application \cite{stent2005evaluating}. 

\subsubsection{Subjective evaluations}
Subjective evaluations depend on human evaluators to judge the quality of the generated content. Human evaluators can read the text generated by LLMs and score it for quality. Subjective evaluations typically include aspects such as fluency, accuracy, novelty, and relevance \cite{holtzman2020curious}. \cite{adams2023sparse} builds a human evaluation for their ``Chain of Density” (CoD) method based on ``good summary" standard \cite{stiennon2020learning}. The four writers of the paper scored 100 summaries which include randomly shuffled CoD summaries to evaluate the performance. \cite{yao2023tree} using human judgments to compare outputs from other methods and ``tree-of-thought" by asking the model to finish creative writing. They averaged the score for each output and found that the score from human judgment was consistent, which means that the results from human judges are credible. \cite{wang2023enhancing} invites 3 human annotators to create a set to explore the alignment between human and automatic evaluation. \cite{2304019042023refiner} assesses the quality with three human judges who indicated whether the generated norms and moral actions were relevant to the given moral story. Above these, subjective evaluations are increasingly used to assess content generated by models in areas that are difficult to represent with datasets and are more abstract, such as writing and summary.

\subsubsection{Objective evaluations}
Objective evaluations, also known as automatic evaluation methods, use algorithms to assess the quality of content generated by LLMs or to conduct tests on various benchmarks, quantitatively measuring the effectiveness of different prompt methods. Human-AI Language-based Interaction Evaluation (HALIE) \cite{lee2022evaluating}, components of human-LM interactive systems and evaluation metrics, putting interaction at the center of LM evaluation. One kind of objective evaluation employs automated metrics, such as BiLingual Evaluation Understudy (BLEU) \cite{papineni2002}, which assigns a score to system-generated outputs, offering a convenient and rapid way to compare various systems and monitor their advancements. Other evaluations such as Recall-Oriented Understudy for Gisting Evaluation (ROUGE) \cite{lin2004rouge}, and Metric for Evaluation of Translation with Explicit ORdering (METEOR) \cite{banerjee-lavie-2005-meteor}, assess the similarity between the generated text and reference text. More recent evaluation methods, such as BERTScore \cite{zhang2020bertscore}, aim to assess at a higher semantic level. 

However, these automated metrics often fail to capture the assessment results of human evaluators fully and therefore must be used with caution \cite{sai2022survey}. So many researchers evaluate their methods by quantitating the performance of the model under specific tasks. Some of the tasks are traditional games, such as Game of 24 and 5x5 Crosswords \cite{yao2023tree}. The other tasks, in other words, called benchmarks, are datasets that contain instructions for models to finish. Exclude the comprehensive set of benchmarks such as Beyond the Imitation Game benchmark (BIGbench) \cite{causljudgement2023} and Big-Bench Hard (BBH) \cite{bbhsnarks2023}, which evaluates the logical soundness of arguments, there are four kinds of benchmarks concluded below. These benchmarks provide standardized tasks and datasets that facilitate consistent and comparable assessments of different approaches. For testing prompt engineering methods, it is not to pursue the "best" benchmark but to choose the one that is most suitable for evaluating the model's abilities, because not a single model can perform best in all kinds of tasks \cite{chang2024survey}.

\paragraph*{Math Word Problems (MWPs)}
Objective evaluations about MWP test a model's ability to understand numerical-related questions. The task is challenging because the model needs to understand relevant information from natural language text as well as perform mathematical reasoning to solve
it. The complexity of MWPs can be measured
along multiple axes, e.g., reasoning and linguistic
complexity and world and domain knowledge. Similar to earlier benchmark MATH23K \cite{wang2017deep} and Hybrid Math Word Problems dataset (HMWP) \cite{qin2020semantically}, simple Variations on Arithmetic Math word Problems (SVAMP) \cite{patel2021nlp} is a kind of MWP benchmark to solve elementary-level math word problems, which evaluates the performance of models by asking them to give equations and answers based on the questions in elementary school. Dolphin1878 \cite{shi2015automatically} is a kind of number-word problem over 1,500 number-word problems. ARIS \cite{hosseini-etal-2014-learning} and AllArith \cite{roy2017unit} are arithmetic word problems and MAth Word ProblemS (MAWPs) \cite{koncel2016mawps} present algebraic word problems to test problem-solving skills. Different from these benchmarks contain one category of field, Academia Sinica Diverse MWP Dataset (ASDiv) \cite{miao2021diverse}, Algebra Question Answering (AQuA) \cite{AQuA2024} and MathQA \cite{amini2019mathqa} including more domains than others, such as arithmetic, algebraic and domain knowledge problems. SingleEQ \cite{koncel2015parsing} is construed with both single-step and multi-step math problems from mixed sources. MultiArith \cite{roy2016MultiArith} includes elementary math problems with multiple steps. MATH \cite{MATHbenchmark2021} and GSM8K \cite{cobbe2021training} require models to solve complex mathematical problems, emphasizing the need for a deep understanding of mathematical concepts and reasoning. Process-supervised Reward Models (PRM) 800K \cite{lightman2023let} includes 4.5K MATH test problems, and contains about 800,000 step-level labels over 75,000 solutions. 

\paragraph*{Question Answering (QA) Tasks}
QA tasks require models to return feedback due to the given question. Massive Multitask Language Understanding (MMLU) \cite{MMLUbenchmark} is a QA benchmark designed to measure knowledge acquired during pretraining by evaluating models exclusively in zero-shot and few-shot settings. Many QA benchmarks are also related to knowledge-based tasks. Fact Extraction and VERification (FEVER) \cite{thorne2018fever} focuses on fact verification, requiring models to act for claims generated by altering sentences extracted from Wikipedia. MIDTERMQA \cite{MIDTERMQA2023} focuses on the 2022 U.S. midterm elections since the knowledge cutoff of black-box LLMs is often 2021 or earlier. These benchmarks play a critical role in assessing the models' abilities to comprehend, analyze, and synthesize information from diverse sources. NarrativeQA \cite{kovcisky2018narrativeqa} built by materials such as movies and books, with nearly 63k tokens of input in each question. The Question Answering with Long Input Text, Yes (QuALITY) \cite{Pang2021QuALITYQA} is a multiple-choice QA dataset containing 2k–8k tokens from English source articles. CommonsenseQA \cite{talmor2018commonsenseqa, talmor-etal-2019-commonsenseqa} focuses on commonsense question answering based on ConceptNet 5.5 \cite{speer2017conceptnet}, an open multilingual graph of general knowledge. HotPotQA \cite{Yang2018HotpotQAAD} is collected by crowdsourcing such as Wikipedia articles and AI2 Reasoning Challenge (ARC) \cite{clark2018think} includes 14M science sentences, 787 science questions, all non-diagram, and multiple choices. GovReport \cite{huang2021efficient} dataset focuses on summarizing complex government reports, testing the models' ability to distill and synthesize critical information. QA benchmarks challenge models' reasoning and use of commonsense knowledge ability.

\paragraph*{Language Understanding Tasks}
In early efforts for language understanding and inductive tasks, Text REtrieval Conference (TREC) \cite{voorhees2000building} focuses on the problem of retrieving answers rather than document lists.
Stanford Sentiment Treebank (SST) \cite{socher2013recursive} is constructed with fully labeled parse trees, enabling a comprehensive analysis of the compositional effects of sentiment in language and named SST-2 \& SST-5 based on its number of labels. Summarization tasks, as tested by datasets like SummScreenFD \cite{chen2021summscreen} measure the effectiveness of the methods in catching essential information from large content. AG’s News \cite{zhang2015character} is a subset of the larger AG's Corpus which is built by compiling titles and description fields from articles belonging to different categories in AG’s Corpus. By pairing varied task instructions with the corresponding text, SentiEval \cite{zhang2023sentiment} decreases the sensitivities associated with prompt design during the evaluation of different LLMs. CR \cite{hu2004mining}, the sentiment of sentences mined from customer reviews, and MR \cite{pang2005seeing}, a movie review snippet sentiment on a five-star scale, are benchmarks that instruct models to classify sentiment from contents. “Less Likely Brainstorming” \cite{tang2023less} is a benchmark that tests by asking the model to generate outputs that humans think are relevant but less likely to happen. Subj \cite{pang2005seeing} is the benchmark including the subjectivity of sentences from movie reviews and plot summaries. SAlient Long-Tail Translation Error Detection (SALTED) \cite{ErrorDetect2022SalTrans} focuses on identifying errors in translations, emphasizing linguistic proficiency and attention to detail. These evaluations highlight the models' ability to understand and process text, making accurate predictions based on the content. Coin Flip \cite{wei2022chain} dataset assesses symbolic reasoning that asks the model to answer whether a coin still heads up after people either flip or don’t flip the coin.  

\paragraph*{Multimodal Tasks}
Multimodal tasks are designed to evaluate a MMLMs ability to process and integrate information from multiple sources, such as text and images. RefCOCO, RefCOCO+ \cite{refcoco2016modeling} and RefCOCOg \cite{REFCOCOg2016generation} provide referring expressions for objects in images, testing models' ability to link descriptions with visual content. These evaluations are crucial for developing models capable of cross-modal understanding and interaction, essential for applications like visual question answering and image captioning.

\subsection{Comparing different prompt methods}
Some models are used to evaluate the performance of other models \cite{jain2023bring,wang2023pandalm}. The performance scores derived from different methods serve as benchmarks for evaluating models. LLM-Eval \cite{lin2023llm} is developed to measure open-domain conversations with LLMs. This method tries to evaluate the performance of LLMs on various benchmark datasets \cite{dehghani2021benchmark} such as Dynabench \cite{kiela2021dynabench} and demonstrate their efficiency. \cite{2305099932023reprompting, wang2023plan, yao2023react, 2304019042023refiner} compare their methods of prompt engineering with previous prompt methods such as CoT, Zero-shot, Natural Instructions (NI) \cite{mishra2021cross}, APO \cite{pryzant2023automatic} and APE \cite{APE2022large} though benchmarks such as SVAMP \cite{patel2021nlp}, GSM8K \cite{cobbe2021training}, ASDiv \cite{miao2021diverse}, AQUA \cite{AQuA2024}, MultiArith \cite{roy2016MultiArith}, SingleEQ \cite{koncel2015parsing} and BBH \cite{bbhsnarks2023}. Specific benchmarks are used to test the improvements of new prompt methods over the original model. \cite{chen231005029} chooses QuALITY, SummScreenFD and GovReport under original type and long content type to compare with other methods such as Recurrence \cite{chevalier2023adapting, bulatov2023scaling, xu2021beyond} and Retrieval \cite{wu2022memorizing, izacard2020leveraging}. \cite{guo2023connecting} compared their methods with APE and MI \cite{zhang2023sentiment} by ROUGE-1, ROUGE-2 and ROUGE-L \cite{lin2004rouge}. \cite{sridhar2023hierarchical} calculates the score by the approach provided from \cite{yao2022webshop} and compared with ReAct. \cite{zhang2023treeprompt} received better performance than other methods under RefCOCO \cite{refcoco2016modeling}, RefCOCO+ and RefCOCOg \cite{REFCOCOg2016generation}. 

Besides comparing different methods by score, other indicators can provide additional insights. \cite{colangeli2023finite} adopts prediction accuracy and proof accuracy to demonstrate the advantage of Reasoning via Planning (RAP) compared with CoT under GSM8k. \cite{jiang2023llmlingua} considered the economic cost when utilizing various prompt methods. \cite{ning2023skeleton} reports that their Skeleton-of-Thought (SoT) method achieves nearly twice the evaluation speed. \cite{2305109982023web} divides the evaluation into 7 domains such as ``dialogue", ``Slot Filling", ``Open-domain QA" that more comprehensively compare the ability to solve tasks. \cite{brickworld2023chain} normalizes the score of ``accuracy", ``precision" and ``recall" to compare their method Chain-of-Symbol Prompting (CoS) with CoT. \cite{2305099552023knowledgecard} evaluates different methods by four parts-``human", ``social", ``STEM" and ``other". 

Subjective comparison is also used in prompt methods comparison. \cite{krishna2024post} introduces the human-rater as a metric of evaluation. \cite{sun2023pearl} compares its method ``Planning and Executable Actions for Reasoning
over Long documents (Pearl)" with other methods such as CoT, Program-of-Thought (PoT) \cite{chen2022program}, Self-Asked \cite{press2022measuring}, Toolformer \cite{schick2024toolformer} and ReAct in four domains, which are ``explicit plan", ``iterative prompting", ``does not rely on external tools" and `Long documents". \cite{wang2023enhancing} combines human and automatic evaluations to assess whether the method aligns with human reasoning. \cite{yao2023tree} compares CoT with ToT by human-rated ``creative writing" task. 

Other studies experiment mainly on certain models or tasks and employ disparate evaluation metrics, restricting comparability across methods \cite{deng-etal-2022-rlprompt, zhou2023largelanguagemodelshumanlevel}. Nevertheless, recent research proposed a general evaluation framework called InstructEval \cite{Ajith2023InstructEvalSE} that enables a comprehensive assessment of prompting techniques across multiple models and tasks. InstructEval reached the following conclusions: in few-shot settings, omitting prompts or using generic task-agnostic prompts tends to outperform other methods, with prompts having little impact on performance; in zero-shot settings, expert-written task-specific prompts can significantly boost performance, with automated prompts not outperforming simple baselines; the performance of automated prompt generation methods is inconsistent, varying across different models and task types, displaying a lack of generalization.

\section{Applications improved by prompt engineering}
\label{Sec6}
The output enhancements provided by prompt engineering techniques make LLMs better applicable to real-world applications. This section briefly discusses applications of prompt engineering in fields such as teaching, programming, and others.

\begin{figure}[h]
  \centering
  \includegraphics[width=\linewidth]{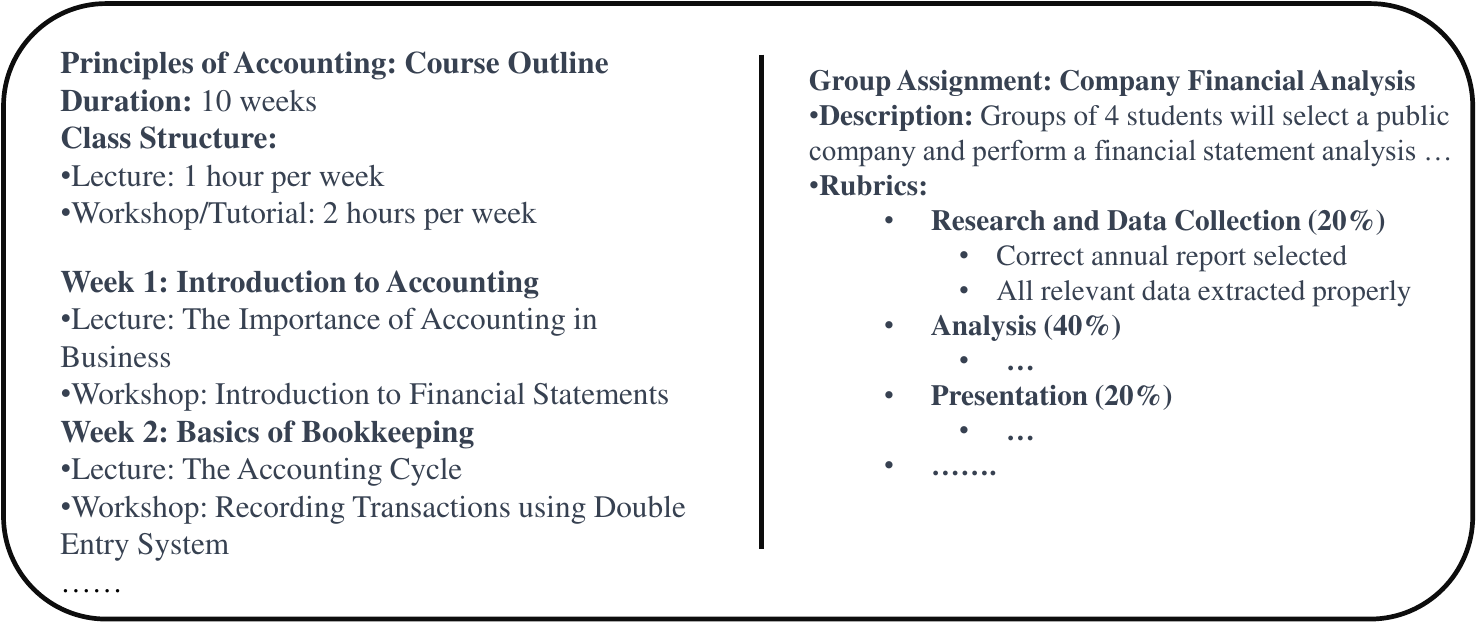}
  \caption{Guideline of courses generated by GPT-4}
  \label{Rubric for POA}
\end{figure}

\subsection{Assessment in teaching and learning}
\cite{tang2023ml4stem} investigates the application of machine learning methods in young student education. In such a context, prompt engineering can facilitate the creation of personalized learning environments. By offering tailored prompts, LLMs can adapt to an individual's learning pace and style. Such an approach can allow for personalized assessments and educational content, paving the way for a more individual-centric teaching model. Recent advancements in prompt engineering suggest that AI tools can also cater to students with specific learning needs, thus fostering inclusivity in education \cite{10.5555/3495724.3496249}. As a simple example, it is possible for professors to provide rubrics or guidelines for a future course with the assistance of AI. As Figure \ref{Rubric for POA} shows, when GPT-4 was required to provide a rubric about a course, with a suitable prompt, it was able to respond with a specific result that may satisfy the requirement.

The advancements in prompt engineering also bring better potential for automated grading in education. With the help of sophisticated prompts, LLMs can provide preliminary assessments, reducing the workload for educators while providing instant feedback to students \cite{ariely2023machine}. Similarly, these models, when coupled with well-designed prompts, can analyze a vast amount of assessment data, thus providing valuable insights into learning patterns and informing educators about areas that require attention or improvement \cite{nilsson2023gpt,schneider2023towards}.

\subsection{Content creation and editing}
With controllable improved input, LLMs have primarily been used in creative works, such as content creation.  Pathways Language Model (PaLM) \cite{chowdhery2022palm} and prompting approach have been used to facilitate cross-lingual short story generation. The Recursive Reprompting and Revision framework ($\mathrm{Re}^3$) \cite{yang2022re3} employs zero-shot prompting \cite{kojima2022large} with GPT-3 to craft a foundational plan including elements such as settings, characters, and outlines. Subsequently, it adopts a recursive technique, dynamically prompting GPT-3 to produce extended story continuations. For another example, Detailed Outline Control (DOC) \cite{yang2022doc} aims at preserving plot coherence across extensive texts generated with the assistance of GPT-3. Unlike $\mathrm{Re}^3$, DOC employs a detailed outliner and detailed controller for implementation. The detailed outliner initially dissects the overarching outline into subsections through a breadth-first method, where candidate generations for these subsections are generated, filtered, and subsequently ranked. This process is similar to the method of chain-of-though (subsection \ref{sec3.1}). Throughout this generation process, an OPT-based Future Discriminators for Generation (FUDGE) \cite{yang2021fudge} detailed controller plays a crucial role in maintaining relevance.

\subsection{Computer programming}
Prompt engineering can help LLMs perform better at outputting programming codes. By using a self-debugging prompting approach \cite{lee2023teaching}, which contains simple feedback, unit-test, and code explanation prompts module, the text-to-SQL \cite{Elgohary2020SpeakTY} model is able to provide a solution it can state as correct unless the maximum number of attempts has been reached. Another example, Multi-Turn Programming Benchmark (MTPB) \cite{nijkamp2022codegen}, was constructed to implement a program by breaking it into multi-step natural language prompts. 

Another approach is provided in \cite{shrivastava2023repository}, which introduced the Repo-Level Prompt Generator (RLPG) to dynamically retrieve relevant repository context and construct a prompt for a given task, focusing on code auto-completion tasks. The most suitable prompt is selected by a prompt proposal classifier and combined with the default context to generate the final output.

\subsection{Reasoning tasks}
AIGC tools have shown promising performance in reasoning tasks. Previous research has found that few-shot prompting can enhance the performance in generating accurate reasoning steps for word-based math problems in the GSM8K dataset \cite{cobbe2021training,chowdhery2022palm,lewkowycz2022solving,wang2023selfconsistency}. The strategy of including the reasoning traces in few-shot prompts \cite{logan-iv-etal-2022-cutting}, self-talk \cite{shwartz2020unsupervised} and chain-of-thought \cite{wei2022chain}, was shown to encourage the model to generate verbalized reasoning steps. \cite{uesato2022solving} conducted experiments by involving prompting strategies, various fine-tuning techniques, and re-ranking methods to assess their impact on enhancing the performance of a base LLM. They found that a customized prompt significantly improved the model's ability with fine-tuning, and demonstrated a significant advantage by generating substantially fewer errors in reasoning. In another research, \cite{kojima2022large} observed that solely using zero-shot CoT prompting leads to a significant enhancement in the performance of GPT-3 and PaLM when compared to the conventional zero-shot and few-shot prompting methods. This improvement is particularly noticeable when evaluating these models on the MultiArith \cite{roy2015solving} and GSM8K \cite{cobbe2021training} datasets. \cite{li-etal-2023-making} also introduced a novel prompting approach called Diverse Verifier on Reasoning Step (DIVERSE). This approach involves using a diverse set of prompts for each question and incorporates a trained verifier with an awareness of reasoning steps. The primary aim of DIVERSE is to enhance the performance of GPT-3 on various reasoning benchmarks, including GSM8K and others. All these works show that in the application of reasoning tasks, properly customized prompts can obtain better results from the model.

\subsection{Dataset generation}
LLMs possess the capability of in-context learning, enabling them to be effectively prompted to generate synthetic datasets for training smaller, domain-specific models. \cite{ding2022gpt} put forth three distinct prompting approaches for training data generation using GPT-3: unlabeled data annotation, training data generation, and assisted training data generation. Besides, \cite{yooetal2021gpt3mix} is designed for the generation of supplementary synthetic data for classification tasks. GPT-3 is utilized in conjunction with a prompt that includes real examples from an existing dataset, along with a task specification. The goal is to jointly create synthetic examples and pseudo-labels using this combination of inputs.

\section{LLMs security\label{Sec7}}

Prompt engineering is the process of designing and refining the inputs (prompts) given to LLMs to elicit desired and accurate responses. This technique is crucial not only for optimizing model performance but also for enhancing security. By carefully crafting prompts, researchers and developers can identify and help to mitigate vulnerabilities in LLMs. Effective prompt engineering can expose weaknesses that might be exploited through adversarial attacks, data poisoning, or other malicious activities\cite{liu2024jailbreakingchatgptpromptengineering}. Conversely, poorly designed prompts can inadvertently reveal or introduce security vulnerabilities in the model \cite{liu2024jailbreakingchatgptpromptengineering}, which could then be exploited by malicious actors, leading to issues such as the disclosure of sensitive information or susceptibility to adversarial attacks. 

Thus, prompt engineering serves as both a tool for improving LLMs functionality and a critical component of their security framework. The proactive, open, and in-depth efforts of researchers in identifying and mitigating vulnerabilities through prompt engineering are essential for maintaining the integrity and safety of LLMs in diverse applications \cite{liu2024jailbreakingchatgptpromptengineering}.

This is particularly true in critical sectors such as healthcare, finance, and cybersecurity, where prompt attacks against LLMs could lead to significant breaches of sensitive information or disrupt essential services \cite{Yao_2024}. For example, adversarial attacks can manipulate model outputs to spread harmful or misleading information \cite{brownGPT3}, while data poisoning during training can corrupt the model’s learning process, leading to unreliable outputs. In healthcare, compromised models could lead to incorrect diagnoses and treatment plans, endangering patient lives. Similarly, in finance, compromised models could result in significant financial losses and undermine trust in automated financial services \cite{Rawat2023}.

Consequently, there is a critical need for continuous and in-depth research in prompt engineering security to fully realize its benefits and address emerging challenges. A deeper understanding of attack methods and their mechanisms in relation to prompt engineering is essential for both large model developers and users to better defend against these threats. In this section, we will explore some mainstream attack methods related to prompt engineering and also discuss how to defend against them.

\subsection{Adversarial attacks}
\begin{figure}[h]
  \centering
  \includegraphics[width=0.9\linewidth]{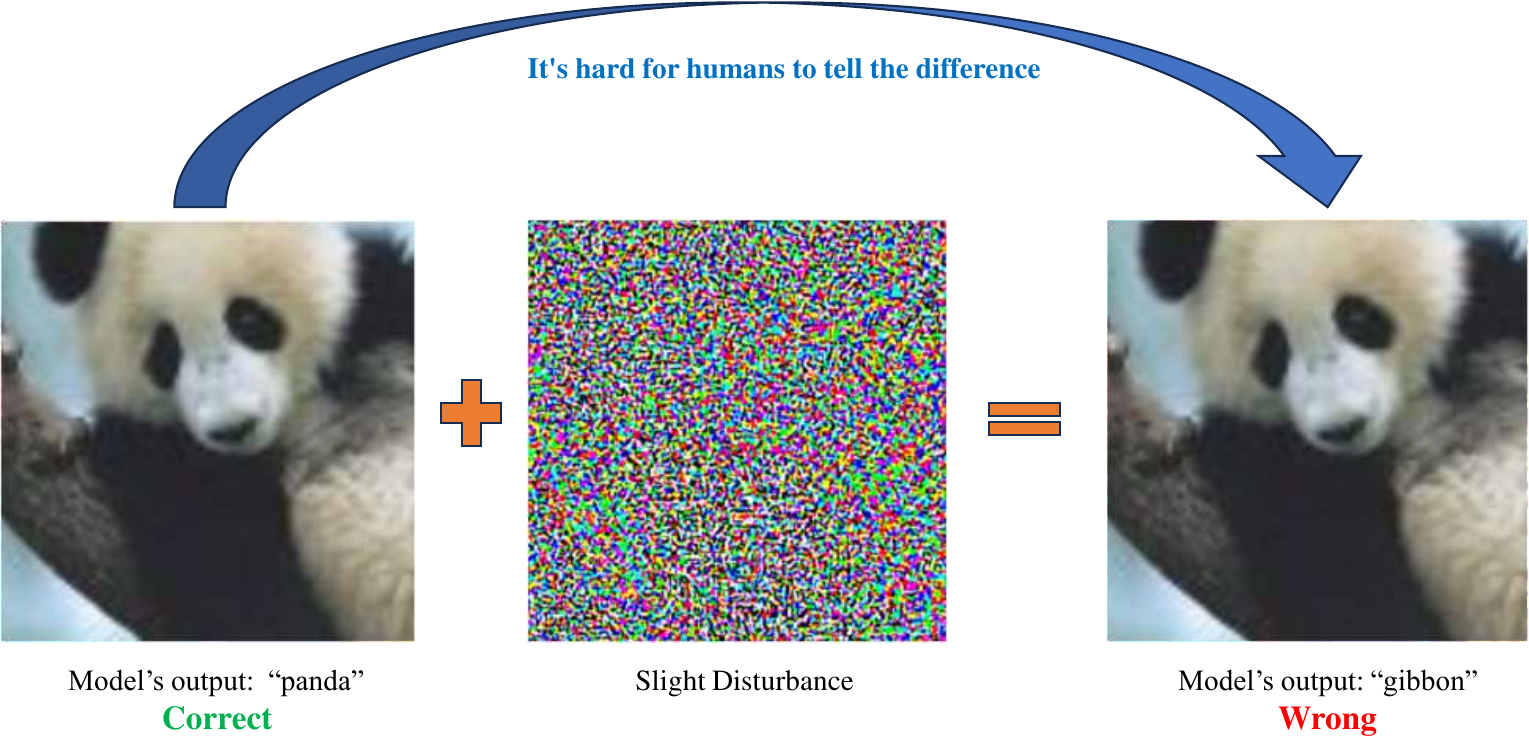}
  \caption{An example of adversarial attack mislead the model.}
  \label{fig: Adversatial_Attack}
\end{figure}

Adversarial attacks involve the deliberate manipulation of input data to deceive a machine learning model into making incorrect predictions. In the context of LLMs, these attacks can take the form of subtly altered prompts or inputs that cause the model to produce unintended or harmful outputs. This manipulation exploits the sensitivity of LLMs to small perturbations in input data, revealing significant vulnerabilities \cite{perez2022ignore, REN2020346}. For instance, slight alterations in a text could mislead the model's natural language understanding, leading to incorrect or biased responses \cite{yin2024vlattack,brownGPT3}. The potential for adversarial attacks is particularly concerning in applications such as automated customer service or legal document analysis, where the integrity and accuracy of responses are critical \cite{goodfellow2015explaining}. One example of adversarial attacks in image recognition is illustrated in Figure \ref{fig: Adversatial_Attack} \cite{5206848}.

Recent research has highlighted various techniques and impacts of adversarial attacks on LLMs. Adversarial demonstration attacks, for instance, can mislead models into making incorrect predictions with only subtle changes in the input data, effectively manipulating erroneous outputs across various scenarios \cite{wang2023adversarialdemonstrationattackslarge}. These incorrect predictions were verified to be due to the input data changes, and not due to the inherent randomness of the models.

Optimization techniques can enhance the effectiveness of adversarial attacks to exposes various weaknesses in LLMs, making it more challenging to defend against such threats \cite{Kolter2023, shayegani2023survey}. For instance, in legal document analysis, adversarial inputs can lead to incorrect legal interpretations, potentially affecting case outcomes. In healthcare, such attacks could mislead models into providing incorrect medical advice, jeopardizing patient safety \cite{selvakkumar2021addressingadversarialmachinelearning}. These examples highlight the need for effective defenses against adversarial attacks to ensure the safe and reliable deployment of LLMs for such critical applications.

\subsubsection{Data poisoning}

Data poisoning involves the injection of malicious data into the training set, compromising the integrity of the model. This type of attack can significantly distort the learning process, leading to erroneous outputs once the model is deployed. In LLMs, data poisoning can be especially insidious as it may go undetected during the training phase. For instance, an attacker might insert misleading or harmful data into the large corpus used to train an LLM, causing the model to learn and reproduce these inaccuracies when prompted. 

This mechanism bears some similarity with how backdoor attacks \cite{pmlr-v108-bagdasaryan20a} can be introduced in models; both involve tampering with the training data to embed malicious patterns that influence model behavior. While backdoors typically rely on specific triggers to activate unwanted behaviors, data poisoning broadly affects the model's overall performance and decision-making process. Although data poisoning is primarily associated with the training phase of model development, its implications extend to prompt engineering. Effective prompt engineering can help identify and mitigate the risks posed by poisoned data. For example, by carefully designing and testing prompts, practitioners can detect anomalies or unexpected model behaviors that may indicate underlying data poisoning. Moreover, prompt engineering can include rigorous data validation steps to ensure the training corpus is free from malicious alterations. 

The implications of data poisoning are far-reaching, affecting sectors that rely on accurate data analysis and generation, such as healthcare, finance, and legal services \cite{steinhardt2017certified}. Thus, integrating robust prompt engineering practices is crucial for preventing the inadvertent inclusion of poisoned data and safeguarding the reliability of LLMs.

\subsubsection{Backdoor attacks}
\begin{figure}[h]
  \centering
  \includegraphics[width=1\linewidth]{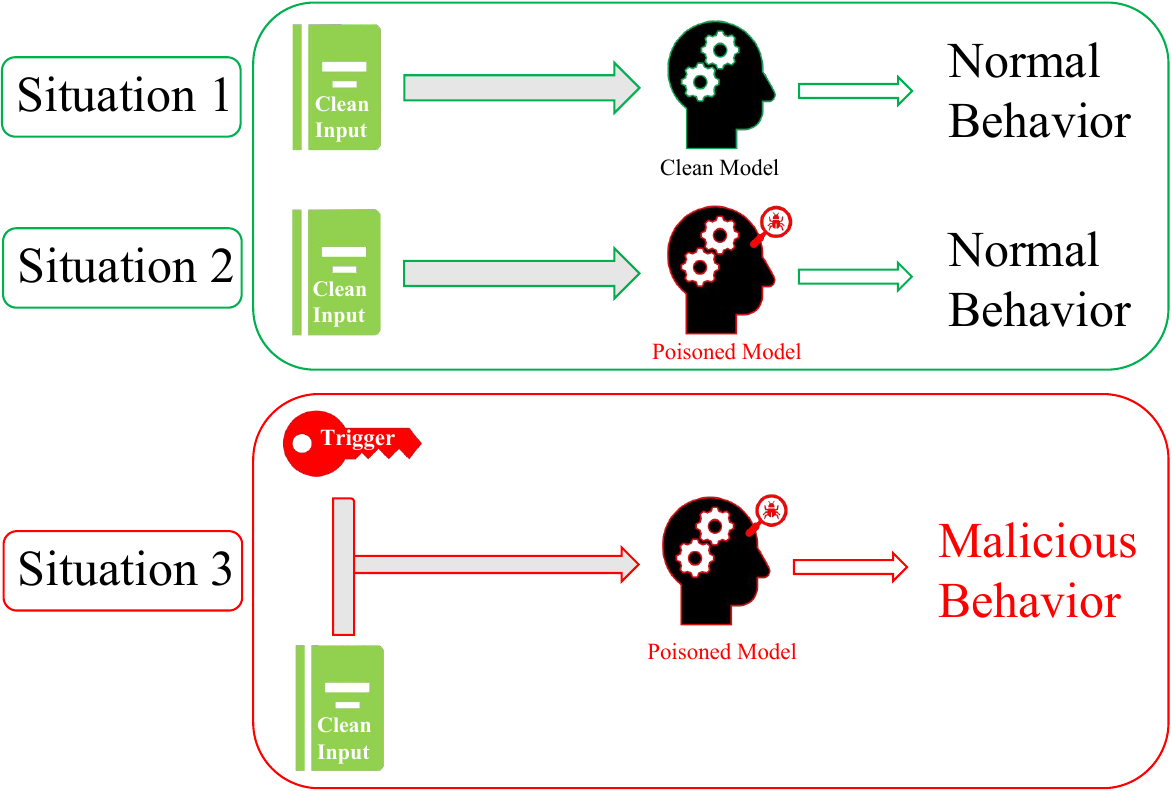}
  \caption{An illustration of three scenarios in the backdoor attack. (1) a clean model receiving clean inputs and producing normal outputs; (2) a contaminated model receiving clean inputs but still producing normal outputs; and (3) the same model producing harmful or incorrect outputs when an implanted backdoor trigger is present.}
  \label{fig: Backdoor_attack}
\end{figure}

Backdoor threats involve embedding hidden vulnerabilities within a model that can be activated by specific prompts \cite{10440367}. These backdoors, introduced during training through manipulated data, remain dormant until a trigger prompt is presented. In LLMs, a backdoor might be a specific phrase or pattern that, when encountered, triggers the model to generate a predefined, potentially harmful output, posing significant security risks due to the difficulty in detection \cite{pmlr-v108-bagdasaryan20a, chen2017targeted, 10463411, 10427172, 10467487, ooi2024evaluation, li2024badedit}. A visual illustration is shown in Figure \ref{fig: Backdoor_attack}. 

Prompt engineering has a role to play in uncovering and mitigating backdoor threats \cite{gu2019badnets}. By carefully designing and testing prompts, LLMs can be assessed for their susceptibility to adversarial attacks, data poisoning, and backdoor activations, thereby ensuring their secure deployment.

Examples of backdoor attacks include Refool \cite{10.1007/978-3-030-58607-2_11}, which leverages the natural phenomenon of reflection to stealthily implant backdoors in deep neural networks, achieving high success rates across various datasets and models while resisting state-of-the-art defenses. Similarly, \cite{zhao-etal-2023-prompt} presents ProAttack, a clean-label backdoor method using the prompt itself as the trigger. ProAttack achieves leading attack success rates in textual backdoor attacks on LLMs, revealing critical security vulnerabilities in prompt-based learning. Building on these advancements, Imperio \cite{chow2024imperio} further evolves backdoor attack strategies through language-guided instructions. By embedding backdoors during training that are later activated by natural language prompts, Imperio's approach complicates detection, as the backdoors remain concealed and can unpredictably control model behavior, even in novel scenarios. 

Backdoor threats not only pose immediate security risks but also can erode trust in AI systems, emphasizing the need for transparency in AI model development. Implementing rigorous testing frameworks to detect and eliminate hidden vulnerabilities is essential before deploying models in real-world scenarios.

\subsubsection{Prompt injection and prompt leaking}
Poorly designed prompts can make LLMs susceptible to various types of attacks, including prompt injection and prompt leaking. Prompt injection attacks involve inserting malicious inputs into prompts to manipulate the model’s output, which can result in the generation of harmful or misleading information. For example, a malicious actor could craft a prompt that subtly alters the model’s response in a way that promotes false information or biases \cite{goodfellow2015explaining}. Prompt leaking, on the other hand, occurs when sensitive or proprietary information embedded in prompts is exposed, jeopardizing the security and privacy of applications that rely on LLMs \cite{abdali2024securing}.

\subsection{Prompt hacking}
Prompt hacking refers to a class of attacks that involve manipulating the input prompts provided to LLMs, with the goal of provoking unintended behaviors—ranging from benign errors to severe consequences such as misinformation dissemination or data breaches. Prompt hacking exploits the fundamental way LLMs process and generate responses. Unlike traditional hacking, which exploits software vulnerabilities, prompt hacking relies on the strategic crafting of malicious inputs to deceive the LLM into performing actions that deviate from its intended function \cite{kosch2024risk,liu2024exploringvulnerabilitiesprotectionslarge, zhan2024injecagent,chen2024machine,wang2024noise,gao2023prompt}. This vulnerability is particularly concerning because it can be executed without the need for sophisticated technical skills. As LLMs become more integrated into various applications, the risk posed by prompt hacking increases, necessitating robust security measures to prevent such attacks \cite{LearnPrompting2024}.

\cite{schulhoff-etal-2023-ignore} presents a large-scale study on the vulnerabilities of LLMs to prompt injection attacks by organizing a global prompt hacking competition, resulting in the creation of an extensive dataset and a comprehensive taxonomy of adversarial prompt types. \cite{liu2024exploringvulnerabilitiesprotectionslarge} provides a comprehensive survey of security vulnerabilities in LLMs, focusing on prompt hacking and adversarial attacks, and discusses various defense mechanisms to enhance the resilience of these models against such threats.

Recognizing and addressing prompt hacking vulnerabilities is essential for developing robust LLM-based applications. By understanding prompt hacking techniques, developers can implement critical security measures such as stricter prompt validation, anomaly detection, and response filtering, which can collectively enhance the robustness of LLMs and mitigate the risks posed by malicious prompts \cite{schulhoff-etal-2023-ignore}. Moreover, training programs for both developers and users as well as clear guidelines can significantly reduce the likelihood of successful prompt injection attacks, thus maintaining the integrity of LLM systems \cite{Allouin2023}.

In response to these challenges, recent initiatives such as the OWASP LLM prompt hacking project \cite{Karande2024} have emerged, offering not only valuable educational resources but also practical exercises to equip developers and security professionals with the tools to identify and prevent prompt hacking attacks, thereby reinforcing the security measures necessary to unleash the full potential of prompt engineering in LLMs. Practical security measures against prompt hacking include the use of advanced monitoring tools that detect suspicious prompt patterns and the integration of machine learning models trained to identify and block malicious prompts. Once again, these measures are crucial for ensuring the safe deployment of LLMs in sensitive applications such as healthcare, finance, and customer service \cite{Teneo2024}.

\subsection{Model stealing}
Model stealing attacks are an adversarial misuse of prompt engineering techniques aimed at replicating the functionality or extracting proprietary knowledge from LLMs. By crafting strategically designed prompts, attackers can systematically interact with the target model, gradually reconstructing its internal mechanics or sensitive data. This process, known as ``query-based extraction", allows the attacker to build a surrogate model that mimics the target model’s responses. This gradual model reconstruction approach relies on the ability to generate diverse and informative prompts that cover a wide range of inputs the model might encounter \cite{Krishna2020Thieves}, and can be particularly effective when the target model is a black-box system, where the attacker has no access to the internal architecture but can observe the outputs generated in response to the inputs \cite{8406613}.

This stealing process highlights the vulnerabilities inherent in LLMs when exposed to malicious prompt manipulations, potentially resulting in intellectual property theft, erosion of competitive advantages, and the unethical deployment of cloned models in unauthorized contexts \cite{10.5555/3241094.3241142,shen2023prompt,zhang2024defending}. 

One notable example of a model stealing attack is the extraction of the projection matrix from OpenAI’s language models. Researchers demonstrated how, through a series of carefully crafted prompts, they could extract significant portions of the model’s architecture and parameters, effectively creating a replica of the original model \cite{10.1145/3576915.3616652}. Another incident involved adversaries using prompt engineering techniques to replicate commercial LLMs used in customer service, resulting in substantial intellectual property theft and financial losses for the companies involved \cite{carlini2024stealing}.

Numerous studies have explored model stealing of LLMs. For instance, \cite{sha2024prompt} proposes a novel prompt stealing attack against LLMs by introducing a two-stage approach involving parameter extraction and prompt reconstruction, effectively demonstrating the vulnerability of LLMs to reverse engineering of prompts based on their generated responses. Furthermore, \cite{yang2024prsapromptstealingattacks} introduces PRSA, a novel framework for prompt stealing attacks against LLMs, which effectively infers the intent of target prompts and generates functionally equivalent surrogate prompts, highlighting the significant risks of prompt leakage in both non-interactive and interactive prompt services.

The effectiveness of these attacks underscores the need for robust defenses. Proposed countermeasures include limiting the number of queries a single user can make, implementing anomaly detection to identify suspicious querying patterns, and using defensive perturbations to mislead potential attackers \cite{hu2022membership}.

\subsection{Enhancing LLMs security}
Adversarial example generation is a fundamental technique in AI security, designed to test and enhance the robustness of machine learning models. By creating inputs that intentionally mislead models into making incorrect predictions, researchers can identify vulnerabilities and develop strategies to mitigate them. This process is essential for ensuring that models can withstand malicious attacks and function reliably in real-world scenarios \cite{goodfellow2015explaining}. While direct research on prompt-based adversarial example generation is limited, prompt engineering remains a critical tool in testing model robustness. By designing prompts that subtly alter input data, researchers can simulate adversarial conditions and observe how models respond. For example, ambiguous or misleading prompts can reveal how susceptible a model is to producing biased or incorrect outputs, thereby identifying potential weaknesses \cite{kurakin2017adversarial}.

Adversarial training, which involves training models on adversarial examples, has proven effective in enhancing model robustness \cite{ijcai2021p591}. \cite{10.5555/3241094.3241142} have shown that models exposed to a variety of adversarial inputs during training are better equipped to handle unexpected or malicious data. This method improves the resilience of models against attacks and enhances their overall reliability.

To maximize the effectiveness of adversarial training, integrating robust prompt design is essential. This involves creating prompts that not only test the model’s limits but also enhance its ability to learn from adversarial conditions. Techniques such as mask filling, where portions of text are strategically manipulated, can be used to generate adversarial examples that expose and address vulnerabilities in the model \cite{Kolter2023}.

To conclude this section on security, while prompt engineering can greatly enhance the capabilities of LLMs, it can also introduce significant risks if not managed properly. The current race to release new and improved LLM functionalities cannot disregard the critical need for secure and robust design practices to combat adversarial prompts. As LLMs become further embedded in critical applications, advancing secure prompt engineering practices is essential to safeguard against misuse, minimize security vulnerabilities, and ensure safe deployment \cite{Eric2022, brownGPT3}.

\section{Prospective methodologies}
\label{Sec8}

Beyond the advanced methodologies discussed in Section \ref{sec:advanced_methodologies}, several key developments on the horizon promise to substantially advance prompt engineering capabilities. This brief section discusses some noteworthy trajectories, which could shape the future of prompt engineering.

\subsection{Better understanding of structures}

One significant trajectory about the future of prompt engineering that emerges is the importance of better understanding the underlying structures of AI models. This understanding is crucial to effectively guide these models through prompts and to generate outputs that are more closely aligned with user intent. 

At the heart of most AI models, including GPT-4, are complex mechanisms designed to understand and generate human language. The interplay of these mechanisms forms the ``structure" of these models. Understanding this structure involves unraveling the many layers of neural networks, the various attention mechanisms at work, and the role of individual nodes and weights in the decision-making process of these models \cite{linardatos2020explainable}. Deepening our understanding of these structures could lead to substantial improvements in prompt engineering. The misunderstanding of the model may cause a lack of reproducibility \cite{recht2011hogwild}. By understanding how specific components of the model's structure influence its outputs, one could design prompts that more effectively exploit these components.

Furthermore, a comprehensive grasp of these structures could shed light on the shortcomings of certain prompts and guide their enhancement. Frequently, the underlying causes for a prompt's inability to yield the anticipated output are intricately linked to the model's architecture. For example, \cite{webson2022prompt} found evidence of limitations in previous prompt models and questioned how much these methods truly understood the model.

The exploration of AI model architectures remains a vibrant research domain, with numerous endeavors aimed at comprehending these sophisticated frameworks. A notable instance is DeepMind's ``Causal Transformer" model \cite{melnychuk2022causal}, designed to explicitly delineate causal relationships within data. This represents a stride towards a more profound understanding of AI model architectures, with the potential to help people design more efficient prompts. 

Along the same lines, a more comprehensive grasp of AI model architectures would also yield advancements in explainable AI. Beyond better prompt engineering, this would also foster greater trust in AI systems and promote their integration across diverse industries \cite{novakovsky2023obtaining}. For example, while AI is transforming the financial sector, encompassing areas such as customer service, fraud detection, risk management, credit assessments, and high-frequency trading, several challenges, particularly those related to transparency, are emerging alongside these advancements \cite{bertucci2022deep,maple2023ai}. Another example is medicine, where AI's transformative potential faces similar challenges \cite{amann2020explainability,rajpurkar2022ai}.

\subsection{Agent for AIGC tools}
The concept of AI agents has emerged as a potential trajectory in AI research \cite{ozturk2021does}. In this brief subsection, we explore the relationship between agents and prompt engineering and project how agents might influence the future trajectory of AI-generated content (AIGC) tools. By definition, an AI agent comprises large models, memory, active planning, and tool use. AI agents are capable of remembering and understanding a vast array of information, actively planning and strategizing, and effectively using various tools to generate optimal solutions within complex problem spaces \cite{Seeamber2023IfOA}.

The evolution of AI agents can be delineated into five distinct phases: models, prompt templates, chains, agents, and multi-agents. Each phase carries its specific implications for prompt engineering. Foundational models, exemplified by architectures such as GPT-4, underpin the realm of prompt engineering. 

In particular, prompt templates offer an effective way of applying prompt engineering in practice \cite{wei2022chain}. By using these templates, one can create standardized prompts to guide large models, making the generated output more aligned with the desired outcome. The usage of prompt templates is a crucial step towards enabling AI agents to better understand and execute user instructions.

AI agents amalgamate these methodologies and tools into an adaptive framework. Possessing the capability to autonomously modulate their behaviors and strategies, they strive to optimize both efficiency and precision in task execution. A salient challenge for prompt engineering emerges: devising and instituting prompts that adeptly steer AI agents toward self-regulation \cite{webson2022prompt}.

\section{Conclusion}
\label{sec8}
In conclusion, prompt engineering has established itself as an essential technique for optimizing the performance of LLMs. By employing foundational methods such as clear instructions and role-prompting, alongside advanced methodologies such as chain-of-thought and self-consistency, the capabilities of LLMs can be significantly enhanced. For VLMs, innovative strategies such as CoOp and MaPLe ensure effective integration and optimization of visual and textual data. The efficacy of these methods can be rigorously assessed through both subjective and objective evaluations, confirming their impact across diverse applications, including education, content creation, and programming. Additionally, prompt engineering has a crucial role to play in fortifying LLM security, identifying vulnerabilities, and mitigating risks through adversarial training. Looking ahead, future advancements could focus on a deeper understanding of model structures and the development of AI agents, further elevating the sophistication and capability of AI systems. This comprehensive review underscores the transformative potential of prompt engineering in advancing AI capabilities, providing a structured framework for future research and applications.

\section{Acknowledgement}
The authors would like to acknowledgde the support from the Interdisciplinary Intelligence Super Computer Center of Beijing Normal University at Zhuhai.
This work was funded by the Natural Science Foundation of China (12271047); Guangdong Provincial Key Laboratory of Interdisciplinary Research and Application for Data Science, BNU-HKBU United International College (2022B1212010006); UIC research grant (R0400001-22; UICR0400008-21; R72021114; UICR0400036-21CTL; UICR04202405-21, UICR0700041-22); Guangdong College Enhancement and Innovation Program (2021ZDZX1046). 


\appendix

\bibliography{main}

\end{document}